\documentclass[final]{article}

\usepackage{neurips_2022}

\usepackage[utf8]{inputenc} 
\usepackage[T1]{fontenc}    
\usepackage{hyperref}       
\usepackage{url}            
\usepackage{booktabs}       
\usepackage{amsfonts}       
\usepackage{nicefrac}       
\usepackage{microtype}      
\usepackage{xcolor}         

\usepackage{graphicx}
\usepackage{caption}
\usepackage{subcaption}
\usepackage{grffile}

\newcommand*\samethanks[1][\value{footnote}]{\footnotemark[#1]}

\title{EN: Vision for Bosnia and Herzegovina in Artificial Intelligence Age: Global Trends, Potential Opportunities, Selected Use-cases and Realistic Goals\\
BOS: Vizija Bosne i Hercegovine u Doba Umjetne Inteligencije: Svjetski Trendovi, Mogućnosti, Odabrani Primjeri i Realistični Ciljevi

}

\author{%
  Zlatan Ajanović\thanks{members of the Association for Advancement Science and Technology (ANNT), Bosnia and Herzegovina.} \\
  Delft University of Technology\\ 
  Delft, The Netherlands\\
  \texttt{z.ajanovic@tudelft.nl} 
  \And
  Emina Aličković\samethanks \\
  Linköping University\\
  Linköping, Sweden
   \And
   Aida Branković\samethanks \\
   CSIRO, Health Intelligence \& \\
   The University of Queensland \\
   Queensland, Australia 
   \AND
   Sead Delalić\samethanks \\
   University of Sarajevo \& Infobip \\
   Sarajevo, Bosnia and Herzegovina 
   \And
   Eldar Kurtić\samethanks \\
   IST Austria \\
   Vienna, Austria 
   \And
   Salem Malikić\samethanks\\
   National Institutes of Health \\
   Bethesda, MD, USA
   \And
   Adnan Mehonić\samethanks \\
   University College London \\
   London, UK 
   \And
   Hamza Merzić\samethanks \\
   Google DeepMind \\
   London, UK 
   \And
   Kenan Šehić\samethanks \\
   Lund University\\
   Lund, Sweden
   \And
   Bahrudin Trbalić\samethanks \\
   Stanford University\\
   Stanford, CA, USA
}

\begin{document}

\maketitle

\begin{abstract}
\textbf{EN:}
Artificial Intelligence (AI) is one of the most promising technologies of the 21. century, with an already noticeable impact on society and the economy. With this work, we provide a short overview of global trends, applications in industry and selected use-cases from our international experience and work in industry and academia. The goal is to present global and regional positive practices and provide an informed opinion on the realistic goals and opportunities for positioning B\&H on the global AI scene.

\textbf{BOS:}
Umjetna inteligencija (eng. Artificial Intelligence, AI) je jedna od najperspektivnijih tehnologija posljednjeg stoljeća koja već uveliko utiče na društvo i ekonomiju. Kroz ovaj rad nudimo kratak osvrt na svjetske trendove, mogućnosti i primjene u industriji te odabrane primjere iz našeg međunarodnog iskustva rada u industriji i akademiji. Cilj je pružiti pozitivne prakse iz svijeta i regiona, te ponuditi stručno mišljenje o realističnim ciljevima i praksama za pozicioniranje BiH na svjetskoj AI sceni.  

\end{abstract}

\section{Uvod}
Kada govorimo o umjetnoj inteligenciji (eng. Artificial Intelligence, AI), prije svega mislimo o razvoju računarskih sistema sposobnih za rješavanje problema koji obično zahtijevaju sposobnost ljudskog rasuđivanja, gdje klasične računarske metode nije moguće praktično primijeniti. U moderno doba, može se reći da interesovanje za AI počinje sa istraživanjima Alana Turinga, pedesetih godina prošloga vijeka. Njen razvoj se eksponencijalno ubrzao u posljednjih par desetljeća ponajviše zahvaljujući razvoju računarske snage i dostupnosti podataka. Sve više digitalnih rješenja koristi neki vid AI, pa tako kada govorimo o AI, ne govorimo više o budućnosti već o sadašnjosti. 

Mogućnosti AI već uveliko prevazilaze naša donedavna očekivanja. Tako smo npr. u 2020. godini svjedočili velikom uspjehu AI u određivanju 3D oblika proteina iz sekvence aminokiselina, a koji predstavlja jedan od najvećih izazova biologije (Jumper et al., 2021). Nadalje, AI model za ljudski govor GPT-3 (Brown et al., 2020) je kreiran da sam piše članke i tekstove visoke kvalitete, a koje nije moguće lahko razlikovati od tekstova napisanih od strane čovjeka. Različiti AI algoritmi specijalizirani za probleme iz domena ljudskog zdravlja su određene medicinske procese dodatno usavršili, učinili ih dostupnim, isplativijim i bržim. Jedan od primjera je algoritam za detekciju karcinoma pluća na osnovu CT slika (Jacobs and van Ginneken, 2019), pomoću kojeg se može postići značajno poboljšanje tačnosti dijagnoze ove bolesti u odnosu na ranije korištene metode. Također, različiti AI algoritmi se koriste u tzv. mozak–računar interfejs (eng. Brain-Computer Interface, BCI). Jedan od takvih primjera je i razvoj algoritama za slušne aparate koji će moći da dekodiraju moždane signale i tako amplifikuju određene govornike i priguše neželjene zvukove (Alickovic et al., 2019).

Tako da bez ikakve rezerve možemo reći da je AI već prisutan u našoj stvarnosti, a mi kao društvo moramo odgovoriti primjereno da bismo omogućili potpunu integraciju AI u korist svih nas i zaštitili se od mogućih zloupotreba. U posljednjih 10 godina, većina naprednih država ulaže velike napore na razvoju i provođenju vlastitih nacionalnih strategija koje za cilj imaju precizno definisanje pozicije AI u svakoj od tih država. Fokus ovih strategija je većinom na rješenjima za povećanje upotrebe AI u privatnom i javnom sektoru kao i dugogodišnjim planovima i investicijama u razvoju same AI. Nadalje, javni i privatni univerziteti su odavno prepoznali potrebu za specijaliziranim bakalaureat i master programima za AI. Privatni sektor je upotrebom AI tehnologija povećao pristupačnost, kvalitetu i sigurnost svojih proizvoda. Stoga države u razvoju, kao što je Bosna i Hercegovina, moraju pronaći vlastiti put o ovom naučno-istraživačkom polju kako bi ostvario puni potencijal. 

Ova studija može koristiti kao početni korak na tom putu. Prema našem saznanju ovo je prva bosansko-hercegovačka AI vizija formalizovana u okviru jednog ovako opširnog rada. Pored opštih informacija na temu AI i pregleda do sada urađenog u nekim od najrazvijenijih zemalja svijeta, studija također obuhvata odabrana iskustva bh. istraživača koji direktno rade na razvoju AI tehnologija kao i njihovu viziju za BiH na polju AI. Cilj nam je ponuditi sveobuhvatan i ažuran pregled trenutnih trendova u AI oblasti uz sugestije kako bh. istraživači u suradnji sa bh. institucijama mogu otpočeti ovaj proces a što bi rezultiralo da BiH iskoristi posljednju priliku i pozicionira se na polju AI. Iako je AI tehnički zahtjevna oblast, rad je napisan kako bi privukao i osobe sa slabijim predznanjem u računarskoj nauci. Pa tako u Poglavlju 2. govorimo o osnovnim pojmovima u AI, primjeni AI u Poglavlju 3. O detaljima utjecaja AI na društvo i ekonomiju govorimo u Poglavlju 4, dok Poglavlje 5 obuhvata detaljniji uvid u odabrana iskustva bh. Istraživača. Stoga Poglavlje 6 sadrži pregleda aktuelnih trendova sa fokusom na nacionalne strategije (Poglavlje 7). Kao glavni doprinos u Poglavlju 8 predstavljamo viziju autora kao i realistične perspektive za Bosnu i Hercegovinu na polju AI.  

\section{Šta je AI?}

AI predstavlja granu nauke koja ima za cilj kreirati programe koji su u stanju replicirati vještine rješavanja problema i proces donošenja odluka prisutan kod čovjeka (Russell and Norvig, 2021). U pojednostavljenom obliku, potrebno je doći do programa koji može što vjerodostojnije imitirati čovjeka. Sam naziv umjetna inteligencija (eng. Artificial Intelligence, AI) je nastao 1956 godine prilikom organizovanja radionice “Dartmouth Summer Research Project on Artificial Intelligence”, bazirano na prijedlogu koji su pripremili John McCarthy, Marvin Minsky, Nathaniel Rochester i Claude Shannon (McCarthy et al., 2006). Kroz godine, nastao je veliki broj definicija umjetne inteligencije, a među neke od najuticajnijih mogu se uvrstiti: John McCarthy-ijeva u radu (McCarthy, 2007) iz 2007. i Alan Turing-ova u radu (Turing, 1950) iz 1950. Pored definicije AI, Turingov rad je ponudio i prijedlog testa, tzv. Turingov test, kojim bi mogli detektovati da li je neki sistem dostigao nivo inteligencije čovjeka. Ovaj poznati test je do danas doživio mnogo modifikacija, ali suština mu se bazira na testiranju da li je čovjek u stanju razlikovati konverzaciju sa AI programom od konverzacije sa drugim čovjekom.

Prema domenu primjenjivosti, umjetna inteligencija se može podijeliti na tri vrste: specijalizirani oblik umjetne inteligencije (eng. Artificial Narrow Intelligence, ANI), generalni oblik umjetne inteligencije (eng. Artificial General Intelligence,- AGI) i nadljudski oblik umjetne inteligencije (eng. Artificial Super Intelligence, ASI). ANI je forma umjetne inteligencije specijalizirana za rješavanje konkretnih zadataka. Komunikaciju sa ovim oblikom inteligencije susrećemo u svakodnevnom životu, kroz interakcije sa pametnim mobitelima, autonomnim vozilima, internet pretraživačima i mnogim drugim savremenim aplikacijama. AGI predstavlja, još uvijek u teoriji, oblik inteligencije koji bi bio svjestan svog postojanja i imao mogućnost rješavanja problema na nivou savremenog čovjeka. I na kraju ASI, također još uvijek u okvirima teorije i filozofskog razmišljanja, bi bio oblik super-inteligencije, inteligencije mnogo superiornije od mogućnosti čovjeka.

\begin{figure}
  \begin{center}
  \includegraphics[width=\linewidth]{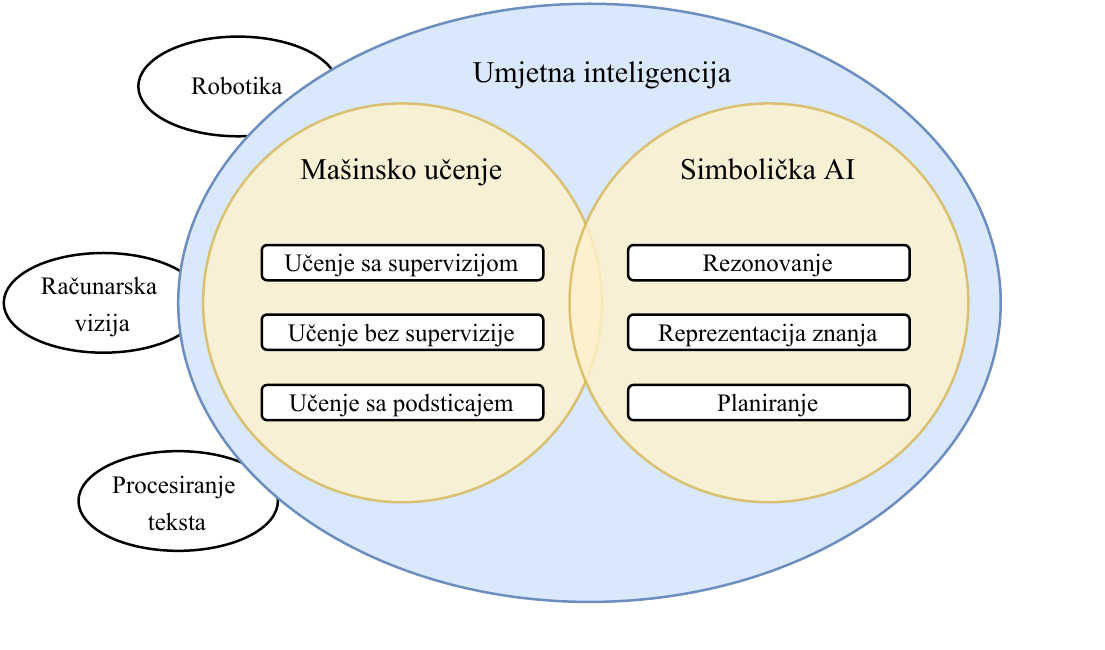}
  \end{center}
  \caption{Pregled AI oblasti.}
  \label{fig:fields}
\end{figure}

Prema pristupu problemu, od samog nastanka polja razlikujemo dvije glavne grane: simbolička i konekcionistička, slika \ref{fig:fields}. 

Mašinsko učenje (eng. machine learning, ML), kao specijalizirani oblik konekcionističke AI, je grana koja je od seminalnog rada 2012. godine (Krizhevsky et al., 2012) pokrenula revoluciju u AI oblasti. U širem smislu, mašinsko učenje se dijeli na tri podoblasti; učenje sa supervizijom (eng. Supervised Learning), učenje bez supervizije (eng. Unsupervised Learning), te učenje sa podsticajem (eng. Reinforcement Learning). Učenje sa supervizijom se bavi funkcijama za modeliranje “ciljanih” vrijednosti, npr. učenje parametara funkcije za predikciju cijena stanova u budućnosti, uzevši u obzir historijske podatke o cijenama stanova iz prošlosti. Cijena stana u ovom slučaju je ciljna vrijednost ili signal za superviziju. Učenje bez supervizije ima za cilj pronalazak strukture u podacima gdje nema konkretnih ciljnih vrijednosti. Tipičan primjer su sistemi za preporuku. Tako npr. Netflix kroz poređenje podataka o filmovima koji su nam se svidjeli sa istim podacima od drugih korisnika može pronaći zajedničku strukturu i time doći do nama relevantnih filmskih preporuka. Učenje sa podsticajem, za razliku od učenja sa i bez supervizije, pretpostavlja interakciju umjetne inteligencije (agenta) sa okolinom. To npr. može biti robot koji usisava prostoriju ili sistem upravljanja procesima fabrike. Cilj agenta je optimizacija nekih od procesa unutar okoline, što u slučaju robota može biti kvalitetnije i brže usisavanje, ili u slučaju fabrike može biti povaćenje efikasnosti (tj. smanjenje troškova) postrojenja.  Mašinsko učenje najčešće koristi modele bazirane na neuronskim mrežama. Takvi modeli, kroz proces treniranja, mogu postići dosta visoku preciznost i određen nivo generalnosti, te prenositi zaključke i na slične slučajeve koje nisu susretali u svojim podacima. Ipak, mašinsko učenje može biti podložno pristrasnosti podataka, slaboj interpretabilnosti, i ponekad može ponuditi rješenja kontradiktorno zdravo-razumskoj logici.

Sa druge strane, simbolička AI (kolokvijalno GOFAI, od engleskog izraza “Good Old-Fashioned AI”) je paradigma koja je bila dominantna od 50ih pa sve do 90ih. Bila je zaslužna za IBM-ov Deep Blue koji je pobijedio tadašnjeg svjetskog šampiona Garry Kasparova (Campbell et al., 2002). Kao prvi simbolički AI sytem smatra se Logički teoretičar (eng. Logic theorist) (Newell and Simon, 1956) koji je bio dizajniran da automatizira ljudsko rezonovanje. Problemi kao što su rezonovanja (eng. reasoning), reprezentacija znanja (eng. knowledge representation) i planiranje (eng. planning) predstavljaju klasične probleme koje rješava simbolička AI. Simbolička AI je bazirana na simboličkom predstavljanju problema, formalnoj logici i metodama pretrage. Simboli su razumljivi ljudima (npr. jabuke i kruške su voće), u kontrast mašinskom učenju koje radi sa vrijednostima razumljivim samo računaru (npr. digitalni zapis slike). Dok se konekcionistička AI fokusira na razvoj AI modela baziranih na podacima, simbolička AI se fokusira na razvoj algoritama koji korištenjem logičkog zaključivanja iz manje podataka mogu donositi nove zaključke. Zbog toga je simbolička AI transparentnija i omogućava generalizaciju ali je također podložnija šumovima, neizvjesnosti, nepotpunom znanju itd.

\section{Primjene AI tehnologije}
Ovaj odjeljak ima za cilj da navede neke od konkretnih primjena baziranih na AI. Zdravstvo: AI je unaprijedila gotovo svaki aspekt zdravstva, od operacija uz pomoć robota do zaštite podataka od sajber (eng. cyber) kriminalaca. Alati bazirani na AI-u smanjuju nepotrebne posjete bolnici, poboljšavaju medicinsku dijagnostiku i liječenje, pomažu u planiranju i rasporedu resursa  bolnica te omogućavaju personaliziranu zdravstvenu njegu. AI je omogućila telemedicinu, te olakšala pružanje zdravstvene zaštite zabačenim i ruralnim djelovima kroz aplikaciju na mobilnom telefonu. Sa strane korisnika olakšan je proces zakazivanja tretmana, plaćanja i liječenja, te zakazivanje termina. Koristeći AI, farmaceutske kompanije istražuju nove lijekove mnogo brže uz smanjene troškove koji su potrebni u tradicionalnom pristupu. Finansije: Efikasnosti i tačnosti AI algoritama je dovela do primjene mašinskog učenja  u financijskim procesima te automatizaciji, chat botovima i algoritamskom trgovanju. Jedan od najvećih finansijskih trendova u 2018. godini je robo-savjetnik, automatizirani portfolio menadžer. Ovi automatizirani savjetnici koriste AI i algoritme za skeniranje podataka na tržištima i predviđanje najboljih dionica ili portfelja na osnovu preferencija. Mnoge banke su već usvojile sisteme zasnovane na AI za pružanje korisničke podrške, otkrivanje anomalija i prevara sa kreditnim karticama. Robotika: Primjeri AI u oblasti robotike su brojni. Današnji roboti na bazi AI pogona, ili mašine koje tako nazivamo, su sposobni rješavati probleme i "razmišljati" u ograničenom kapacitetu. Inteligencija ovih sistema je posljedica procesa učenja kroz simbiozu različitih naučnih disciplina u kombinaciji sa savremenim algoritmima AI. Primjer jednog kompleksnog sistema je humanoidni robot Sofija baziran na AI koji može komunicirati i izražavati facijalne ekspresije da prenese emocije slične čovjeku. Samoupravljajuća vozila: Opremljeni sa velikim brojem senzora koji prikupljaju sve što se događa oko i u vozilu, ova vozila koriste AI da analiziraju podatke i reaguju adekvatno u milisekundama.  Društvene mreže i online platforme: AI je u srži društvenih platformi poput Twittera, Facebooka, Snapchata, itd. AI algoritmi predlažu ljudima da prate i dijele vijesti na osnovu individualnih preferencija korisnika, za praćenje i kategorizaciju video sadržaja na osnovu predmeta, prepoznavanje govora mržnje i terorističkog jezika, i sl. Pored toga, AI se uveliko koristi na online platformama za prodaju i kupovinu (e-commerce) te online platformama za gledanje sadržaja (engl. streaming) gdje se na osnovu prethodnih isustava predlažu sadržaji i proizvodi koji bi mogli zanimati korisnika (engl. recommendation). Turizam: AI je postala neizostavni dio transportne i turističke industrije. AI algoritmi za preporuku se koriste za pronalazak najboljih mogučnosti reflektoiranih kroz cijenu i druge spcifikacije korisnika. Istraživanje svemira: AI je našla primjenu u izučavanju ogromnih količina podataka prikupljenih sa Kepler teleskopom i kompjuterskoj viziji NASA-inog roverera na Marsu. Igrice: AlphaGo i F.E.A.R softveri su primjeri u kojima je AI je postala sastavni dio industrije igara kroz igru između čovjeka i mašine kao oponenta. Proračuni i simulacije: Kombinovanje AI sa računarskom mehanikom fluida (Raissi et al., 2017) omogućilo je efikasno upravljanje i održavanje kompleksnih sistema kao što su avioni ili nuklearna postrojenja gdje je izuzetno bitno adekvatno kvantificirati svaku potencijalnu nesigurnost. Neuroznanosti (eng. neuroscience) i neurotehnologija (eng. neurotechnology): Kombinovanjem AI-a sa  BCI sistemima je omogućila dekodiranje mozga i tako direktnu komunikaciju između čovjekovog mozga i vanjskih uređaja (npr. računar, slušni aparati, neuroprostetski pomagala, itd.).

\section{Utjecaj AI tehnologija na društvo i ekonomiju}
Kao što se često kolokvijalno predstavlja, može se reći da AI simbolizira električnu energiju 21. stoljeća (Ng, 2018). Kao što je električna energija oblikovala i unaprijedila svijet u 19. stoljeću, očekuje se da AI oblikuje svijet u 21. stoljeću. Napredak AI-a doveo je do niza promjena u svim domenama ljudskog djelovanja. Razvoj inteligentnih rješenja doveo je do automatizacije velikog broja procesa, ponudio je rješenja koja sustižu ili prevazilaze ljudske mogućnosti. To je dovelo do promjene ekonomskih mogućnosti, ubrzanja procesa, proizvodnje i ušteda u poslovanju. Kreirane su nove mogućnosti za zaradu, ali su i neke grane industrije u potpunosti promijenjene i prilagođene novim uslovima.

AI i razvoj online platformi promijenili su niz industrija, poput filmske ili trgovačke grane(npr. primjenom sistema preporuka na online platformama). Osim sistema preporuka, generisan je niz drugih personaliziranih sadržaja i usluga, pri čemu za personalizaciju usluge nisu potrebni dodatni ljudski resursi, što čitav proces čini efikasnijim. Osim govornih asistenata (npr. Siri, OK Google, Bixby, Cortana), mobilni uređaji su opremljeni nizom drugih pametnih algoritama koji olakšavaju i čine svakodnevnicu učinkovitijom. 

AI tehnologija je pronašla niz korisnih primjena u razvoju nauke. Medicina je značajno unaprijeđena primjenom AI tehnologija. Osim procesa kreiranja lijekova, uspostavljanja jednostavnih dijagnoza i terapija, AI pomaže u kreiranju ortopedskih i neuroprostetskih pomagala ili otkrivanju teških oboljenja (Basu et al., 2020). U (The Royal Society, 2019), opisan je niz primjena u raznim oblastima, poput otkrivanja strukture proteina, razumijevanja klimatskih promjena, orktivanju uzoraka i novih tijela u astronomiji, ali i niz drugih primjena u hemiji, fizici, biologiji i drugim naukama.

Osim pozitivnih uticaja na društvo, AI ima nekoliko nedostataka koji se trebaju prevazići. Kako je ranije navedeno, AI optimizuje procese i resurse. Time se smanjuje potreba za ljudskom random snagom, što se posebno izražava u repetitivnim poslovima. Po istraživanju PwC, više od 7 miliona radnih mjesta će biti zamijenjeno AI rješenjima do 2037. godine u Velikoj Britaniji. Međutim, očekivano je otvaranje još većeg broja novih radnih mjesta (7.2 miliona), pri čemu se repetitivni poslovi mijenjaju poslovima koji zahtijevaju kreativnost i u kojima ljudi imaju prednost u odnosu na  AI (PricewaterhouseCoopers, 2018).

AI algoritmi su prvenstveno zasnovani na analizi i upotrebi velikih količina podataka. Mogućnost primjene modernih AI tehnika raste povećanjem količine podataka, te je manja mogućnost pronalaska kvalitetnih rješenja bez velike količine podataka. Time se smanjuje mogućnost uspjeha manjim kompanijama, te dolazi do centralizacije moći. Istovremeno, ključno pitanje u prikupljanju podataka je privatnost. Ideja Federativnog učenja (eng. Federated learning, FL) nudi mogućnost kolaborativnog pripremanja modela. Svaka stranka koja učestvuje u treniranju ima mogućnost čuvanja svojih podataka, a radi se samo modifikacija samog modela i njegovih parametara. Pri tome, ne dolazi do direktne podjele podataka i čuva se privatnost korisnika.

\section{Odabrani primjeri iz iskustva bosansko-hercegovačkih istraživača}
Iako je razvoj AI u BiH na nezavidnom nivou, pojedini bh. istraživači već uveliko prave značajne pomake na polju AI-a kroz rad kako u Bosni i Hercegovini tako i u inostranstvu. Iskustvo takvih pojedinaca je od neprocjenjive važnosti za društvo. Stoga u ovom poglavlju govorimo o manjem broju odabranih primjerima kako bi približili javnosti savremena AI istraživanja od strane bh. istraživača. Sam odabir je prvenstveno prema iskustvima autora i ne predstavlja sveobuhvatni pregled cjelokupnog istraživanja bh. istraživača.

\subsection{Digitalizacija urgentne medicine i akutne njege}
Tranzicija bolnica ka potpuno digitalnom okruženju na globalnom nivou dovela je do revolucionarne promjene u načinu pružanja zdravstvene zaštite i praćenje pacijenata. Prikupljanje podataka o pacijentima u elektronskim medicinskim kartonima (eng. Electronic Medical Records; EMR) otvorilo je priliku za učinkovitu, preciznu i personaliziranu zdravstvenu njegu. Jedan od primjera upotrebe digitalnih alata razvijenih koji koriste EMR u realnom vremenu su alati za praćenje vitalnih parametara i algoritmi za automatsku detekciju kliničkog pogoršanja pacijenata u akutnoj njezi vezanih za a neželjenim događajima kao što su neplanirani transfer na intenzivnu njege, srčani udar, smrt, sepsa. Drugi primjer je prevencija neplaniranih hospitalizacija, uključujući ponovni prijem i  javljanje pacijenata na odjeljenje hitne pomoći koji je bitan za adekvatno rješavanje problema bolničkih kapaciteta i rastuće potrebe za bolničkom njegom. Treći primjer je razvoj digitalnih alata za početne trijaže i stratifikacije rizika primljenog pacijenta. Primjena AI-a u urgentnoj radiologiji je jedna od najviše izučavanih oblasti. Naime, doktori urgentne medicine često imaju zadatak brze identifikacije patologija koje ugržavaju život i poduzimanju akcije skladno rezultatima snimaka prije nego što snimak bude pregledan od strane radiologa. 
Konvencionalni tokovi rada i detekcije ranog pogoršanja stanja pacijenta oslanjaju se uglavnom na kliničko bodovanje i alate za praćenje i okidanje alarma zasnovanih na određenim pravilima. Pokušaj da se adresira nedostatak metodološke strogosti postojećih metoda dovela je do razvoja i implementacije inteligentnih algoritama koji imaju za cilj da predikciju pogoršanja stanja pacijenta (Shin et al. 2021, Markazi-Moghaddam, Fathi, and Ramezankhani 2020, Brankovic et al. 2022b) i ili čak korak ispred - predikciju aktivacije alarma za detekciju ranog pogoršanja stanja pacijenta (Brankovic et al. 2022a). 

Iako se može naći velik broj literature i predloženih inteligentnih modela, vrlo je malo literature (samo 146 u periodu od 2004 do 2021) koja daje uvid u rezultate registriranih kliničkih studija primjene razvijenih prediktivnih modela  na bazi AI u urgentnoj medicine i odjeljenju intenzivne njege (Liu et al. 2021), pri čemu je većina prediktivnih modela u formi linearne regresije. Jedan od glavnih razloga za spor ulazak kompleksnijih modela poput onih baziranih na dubokim neuronskim mrežama jeste kad su u pitanju digitalni uređaji za predikciju jeste netransparentnost ovih modela i samim tim rezistivnost kliničkog osoblja da se pouzda u predviđanje bez rezonovanja i shvatanja principa po kojem je digitalni alat napravio prognozu. Također, kompjutaciona kompleksnost i implemetacija ovih digitaklnih alata u klinički tok rada u realnom vremenu je također jedan od izazova. U konačnici malo ili nikakvo poboljšanje (Christodoulou et al. 2019) kompleksnih prediktivnih inteligentnih modela spram modela baziranih na regresiji u određenim kliničkim aplikacijama (Christodoulou et al. 2019) također su neki od razloga za spor prodor i simbiozu kompleksnih algoritama mašinskog ućenja i AI sa kliničkom sredinom. Nešto manja strogost u pogledu zahtjeva za transparentnosti modela je prisutna u radiologiji, ali isto tako i velika tačnost u uočavanju uzoraka digitalnih uređaja baziranim na umjetnim neuronskim mrežama (Goldstein et al. 2017). Jedno od mogućih objašnjenja je sama primjena gdje je algoritam odnosno digitalni alat u osnovi komplement i pomoć stručnjaku koji i dalje ima glavnu ulogu u odlučivanju (Prevedello et al. 2017).  

\subsection{Čovjek-računar interfejs}
Čovjek-računar interfejs (eng. brain-computer interface) je termin koji opisuje direktnu komunikaciju između čovjekovog mozga i vanjskih uređaja kao što su računari, slušni aparat ili neka druga neuro-prostetska pomagala. Neuronski signali u ljudskom mozgu se mogu mjeriti invazivnim metodama koje koristi elektrode stavljene direktno na površinu mozga (npr. elektrokortikografija - ECoG ili intrakranijalna elektroencefalografija - iEEG) ili neinvazivnim metodama koje koriste elektrode koje se stavljaju na kožu glave (npr. elektroencefalografija - EEG) ili u uši (npr. earEEG). Kombinovanjem AI algoritama i BCI sistema moguće je dekodirati i klasificirati različita ljudska stanja, npr. koga slušamo (Alickovic et al., 2019), koga gledamo (Bilert et al., 2020), kakvog smo raspoloženje (Sani et al., 2018), neurološka oboljenja (Alickovic et al., 2018b), (Alickovic and Subasi, 2018a), pokrete dijelova našeg tijela (Alickovic and Subasi, 2018a), itd. 

Jedna od značajnijih primjena kombinovanja AI tehnologija sa BCI sistemima je u razvoju nove generacije slušnih aparata i pomagala. Nove generacije slušnih aparata bi trebala imati mogućnost oponašanja ljudskog mozga koji ima izvanrednu sposobnost odabira i amplifikacije pojedinih govornika i prigušenja neželjenih zvukova u bučnom srednima, kao što je prepun kafić ili školsko dvorište. Trenutno ni najsavremeniji slušni aparati nemaju tu sposobnost. Veliki napredak u ovom polju je već učinjen, gdje je dokazano da se moždani signali mogu koristiti da se dekodira auditornja pažnja, tj.amplifikacija pojedinačnih govornika na osnovu nuronskih signala (Alickovic et al., 2019), (Alickovic et al., 2020), (Alickovic et al., 2021), (Andersen et al., 2021), (Geirnaert et al., 2021). Kreiranjem ovakvih slušnih aparata i pomagala koja koriste moć samog mozga (tj. neuronske signale), dovesti će se do tehnoloških poboljšanja koja će stotinama milijuna osoba oštećenog sluha diljem svijeta  (predviđa se otprilike oko 900 miliona ljudi u 2025. godini i oko 2.5 milijarde ljudi u 2050. godini s oštećenjem sluha) omogućiti da komuniciraju sa istom lahkoćom kao i njihovi prijatelji i porodica.

\subsection{Rano prepoznavanje karcinoma}
Karcinom je danas jedan od vodećih uzročnika smrti u cijelom svijetu. Procjenjuje se da je samo u 2020. godini u svijetu dijagnosticirano oko 19.3 miliona novih slučajeva oboljenja od karcinoma, dok se broj smrtnih slučajeva prouzrokovanih ovom grupom bolesti u istoj godini procjenjuje na oko 10 miliona (Sung et al., 2021). Osim fizičke boli, emocionalnog stresa i svih drugih poteškoća s kojima se susreću i bore pacijenti i njihove porodice, karcinom također predstavlja i veliki teret socijalnim i zdravstvenim sistemima. S ciljem smanjenja incidentnosti i poboljšanja kvaliteta života pacijenata, države širom svijeta danas ulažu ogromne resurse u istraživanja karcinoma. Ova istraživanja, između ostalog, za cilj imaju razumijevanje njegovog nastanka i razvoja, dizajn metoda za ranu detekciju i monitoring, kao i razvoj novih i poboljšanje već postojećih tretmana liječenja. U ovom kontekstu, umjetna inteligencija ima širok spektar primjena, a u ovom radu ćemo se fokusirati na dva konkretna problema u kojima je njena upotreba do sada dala izuzetno obećavajuće rezultate. 

Prva, i do sada jedna od najuspješnijih, primjena umjetne inteligencije u istraživanju karcinoma vezana je za analizu histopatoloških slika. Algoritmi bazirani na deep learning-u  su se pokazali izuzetno uspješnim kako u detekciji malignih oboljenja iz ove vrste podataka, tako i u određivanju nekih važnih kliničkih parametara poput vrste karcinoma, njegovog stadija i procenta kancerogenih ćelija u tkivu (Echle et al., 2021). Jedno od značajnijih istraživanja iz ovog domena je istraživanje objavljeno 2017. godine (Esteva et al., 2017), u kom je predstavljen algoritam za detekciju karcinoma kože baziran na korištenju deep convolutional neural networks. Nakon što je treniran na skupu podataka od 129450 slika, algoritam je testiran na kolekciji slika za koje je (sa velikom pouzdanošću) poznata tačna klinička dijagnoza (maligno vs. benigno). Rezultati su potom poređeni sa predikcijama načinjenim na istom skupu slika od strane 21 certificiranog dermatologa. U konačnici se pokazalo da nije bilo značajne razlike u tačnosti između dvije grupe predikcija, tj. rezultati dobijeni koristeći algoritam parirali su onim dobijenim direktnim pregledom od strane specijalista. Potencijal algoritama u automatskom prepoznavanju malignih promjena na koži, kao i široka rasprostranjenost pametnih telefona koji omogućavaju uzimanje/dobijanje slika na jednostavan način, podstakli su i razvoj sofisticiranih aplikacija namijenjenih za analizu ovako dobijenih slika. Pretpostavlja se da će ove aplikacije u budućnosti uveliko pomoći ranoj detekciji karcinoma kože, kada je bolest lokalizovana i u mnogim slučajevima ju je moguće u potpunosti odstraniti hirurškim putem. Pored primjena u analizi slika kože, algoritmi zasnovani na metodama umjetne inteligencije su se također pokazali veoma korisnim i u analizi histopatoloških slika karcinoma dojke (Jannesari et al., 2018), prostate (Tolkach et al., 2020), pluća (Yang et al., 2021) i drugih. 

Druga primjena na koju ćemo se fokusirati u ovom članku je detekcija mutacija (promjena) na DNK kodu iz uzoraka krvi. DNK mutacije se danas smatraju jednim od glavnih uzročnika nastanka i razvoja karcinoma te stoga njihova rana detekcija može biti od neprocjenjive koristi za ranu detekciju bolesti. Većina mutacija značajnih za karcinom nastaju tokom života (tj. nisu nasljedne), ali ih je zbog tehnoloških ograničenja nemoguće otkriti prilikom standardnih ljekarskih pregleda. Zahvaljujući napretku tehnologija za sekvenciranje DNK i razvoju kompjuterskih algoritama za analizu podataka generisanih ovim tehnologijama, odnedavno je moguće odrediti prisutnost određenih mutacija u organizmu analizom laboratorijskih uzoraka krvi. Naime, odumiranje starih i nastanak novih ćelija su svakodnevni procesi u našem organizmu. Prilikom odumiranja ćelije, mali fragmenti (dijelovi) njenog DNK završavaju u krvi u kojoj cirkulišu određeno vrijeme. Odumiranjem ćelija sa mutiranim DNK u krv ulaze njihovi DNK fragmenti koji sadrže signale za prisutnost mutacija. Ovi signali se potom prilikom sekvenciranja prenose u skup generisanih podataka koji se zatim prosljeđuju na kompjutersku analizu. Poznato je da detekcija mutacija prilikom kompjuterske analize nije nimalo jednostavan zadatak. Jedan od glavnih razloga za to je činjenica da veliku većinu DNK fragmenata u krvi čine fragmenti zdravih ćelija koji ne sadrže mutacije, što je pogotovo izraženo u ranoj fazi razvoja karcinoma. Pored toga, tokom samog procesa DNK sekvenciranja dolazi do grešaka u učitavanju DNK koje mogu dati lažni signal za postojanje mutacije tamo gdje ona u stvarnosti ne postoji. Stoga se u praksi susrećemo sa tkzv. "low signal-noise ratio" problemom. Neka od prvih rješenja ovog problema su bazirana na upotrebi specifičnih distribucija pomoću kojih se određuje vjerovatnoća postojanja mutacije (Kockan et al., 2017), dok se najnoviji metodi oslanjaju na statističko modeliranje i upotrebu mašinskog učenja (Li et al., 2021). Naprednija rješenja ovog problema su ponuđena od strane biotehnoloških kompanija koje posjeduju velike skupove kvalitetnih podataka koje koriste za razvoj i treniranje algoritama umjetne inteligencije, ali se njihovi detalji obično čuvaju kao poslovne tajne i nedostupni su široj akademskoj zajednici (“SOPHiA artificial intelligence empowers liquid biopsies to fight cancer,” 2017). Nedavno su također predložena i neka rješenja bazirana na metodama mašinskog učenja koja pomažu pri samom dizajnu panela za DNK sekvenciranje (Cario et al., 2020) što za krajnji rezultat može imati poboljšan nivo detekcije mutacija.

\subsection{Savijanje proteina}
Problem savijanja proteina (Dill et al., 2008) je jedan od ključnih problema biologije u posljednjih 50 godina. Proteini su kompleksne molekule sastavljene od niza aminokiselina. Funkcija proteina u najvišoj mjeri zavisi od njegove unikatne 3D strukture - transport nutrienata, pospješenje hemijskih reakcije, izgradnja tkiva, uloga antitijela, samo su neke od korisnih uloga proteina u organizmu. Predikcija 3D strukture proteina na osnovu odgovarajućeg lanca aminokiselina je problem “savijanja proteina”. Rješenje ovog problema otvara vrata nekim od trenutno najvećih izazova čovječanstva kao što je pronalazak novih lijekova, razgradnja industrijskog otpada, te mnogi drugi. 

Težinu ovog problema je teško objasniti u okviru ovog rada, ali činjenica je da se cijeli doktorski i postdoktorski studiji nerijetko bave rješenjem problema savijanja samo jednog ili vrlo malog broj proteina. Uz mukotrpan višegodišnji posao za otkrivanje jedne strukture, potrebno je imati i posebne višemilionske uređaje koji mogu mjeriti udaljenosti na nivou atoma (ispod jednog nanometra). Nakon više desetljeća rada na ovom problemu, otkrivena je struktura svega 17\% od oko 20 hiljada proteina u ljudskom tijelu, odnosno 170 hiljada od ukupno preko 100 miliona svih nama poznatih proteina.

Da bi se ubrzao progres u ovom polju, 1994. pokrenuto je CASP takmičenje (Moult, 2005) koje svake dvije godine evaluira metode za predikciju proteinskih struktura. Evaluacije se vrši na neobjavljenim strukturama otkrivenim u prethodne dvije godine. DeepMind je po prvi put učestvovao u CASP13 takmičenju 2018. godine i koristeći metode mašinskog učenja sa i bez supervizije ostvario ogroman pomak u odnosu na već dostupne metode. Već 2020. godine na CASP14 takmičenju, AlphaFold tim (Jumper et al., 2021) iz DeepMind-a je uspio rafinirati prethodne modele i dobiti rezultate sa greškom koja parira eksperimentalnim rezultatima na skupocjenim uređajima - drugim riječima, problem savijanja proteina je riješen! 2021. godinu AlphaFold tim je proveo radeći na dodatnom rafiniranju modela, objavljivanju koda za širu upotrebu, te rješavanju problema savijanja proteina prvo za sve proteine u ljudskom tijelu, pa onda i za sve nama poznate proteine.

Iako poznajemo strukture pojedinačnih proteina, ostaju otvorena pitanja interakcije istih sa drugim molekulama, te način na koji se dva ili više proteina povezuju u kompleksnije strukture. Nevjerovatno je da se jedno ovakvo otkriće, koje ima potencijal da revolucionira tok moderne biologije, farmacije, i medicine, desilo tik na pragu ove godine. I ne sumnjamo da će napredne tehnike mašinskog učenja korištene za realizaciju ovog projekta biti od neprocjenjive koristi za nadolazeće probleme. 

\subsection{Planiranje i učenje u robotici}
Broj industrijskih robota u fabrikama širom svijeta dosegao je čak 2.7 milliona (uključujući robote dostavljene završno sa 2019. godinom) (IFR, 2020). U ovaj broj nisu uključeni kućni roboti kao sto su roboti usisivači, kosilice, itd. Iako su izuzetno korisni i uspješni  u obavljanju poslova za koje su dizajnirani, svi pomenuti roboti su ipak daleko od nivoa inteligencije koja se očekuje od inteligentnog robota. Roboti u industriji su programirani za tačno određeni zadatak (npr. varenje automobilske školjke na tačno definisanim tačkama) i teško se prilagođavaju čak i na male promjene u svom okruženju. Zbog toga, kao i zbog sigurnosti ljudi (sa kojima još uvijek ne znaju da sarađuju), roboti se drže u fizički odvojenim i kontrolisanim prostorima. Slična situacija je i sa kućnim robotima koji rade na principu jednostavnih pravila te često zahtijevaju određene prilagodbe u kućanstvu da bi mogli uspješno obavljati svoj zadatak (npr. usisavanje). Prema tome, vidimo da nivo inteligencije robota trenutno dozvoljava robotima obavljanje samo limitiranih, tačno definisanih zadataka u okruženju dizajniranom za robota.

Da bi dosegli punu iskorištenost, roboti moraju naučiti da operiraju u okruženju dizajniranom za čovjeka, u saradnju sa čovjekom te da se prilagodjavaju novim zadacima.  Zbog toga, trenutno istraživanje i razvoj se bavi takozvanim kolaborativnim robotima (engl. Cobot) koji mogu raditi u prisustvu čovjeka. Već postoje roboti koji su hardverski razvijeni namjenski kao Coboti i bezbjedniji su za upotrebu u prisustvu čovjeka. Ipak softver (inteligencija) tih robota još uvijek nije na zavidnom nivou te se kroz istraživanje teži razviti robote koji mogu da uče od ne-eksperta ljudi i koji mogu da se prilagođavanju na svakodnevne zadatke. Posebno važan za ovo je problem planiranja zadataka i kretanja (engl. Task and Motion Planning, TAMP) koji teži, da na osnovu zadanog krajnjeg cilja, robotu pruži listu svih potrebnih zadataka (međukoraka) i kretanje za izvršavanje tih zadataka (Garrett et al., 2021). Praktičan primjer je npr. robot koji, radi izvršenja cilja da korisniku doda mlijeko, autonomno odlučuje da otvori frizider, izvadi mlijeko i zatvori frižider ili autonomno vozilo koje, kada vozi u urbanom saobračaju, može autonomno odlučiti da npr. treba da ubrza, prebaci se u lijevu traku, pretekne drugo vozilo i vrati se u prvobitnu traku (Ajanovic et al., 2018) ili da bi vozilo maksimalnom brzinom na putu sa malim trenjem, može da autonomno odlučuje kako treba da priđe krivini, gdje i koliko da izgubi djelimično kontrolu nad vozilom uz drift (po uzoru na ekspert reli vozače) te kako da izađe iz krivine sa maksimalnom brzinom (Ajanovic et al., 2020, Ajanovic et al., 2022). Ovi problemi su izrazito računarski zahtjevni te zahtjevaju modele koji nisu uvijek dostupni (Ajanovic, 2019). Stoga se pribjegava korištenju metoda učenja koje nadomještaju to.

Učenje u robotici nije jednostavno jer konvencionalni pristupi učenja (npr. Reinforcement Learning (Kober et al., 2013)) zahtjevaju dosta iteracija pokušaj-pogreška koji se ne mogu priuštiti u stvarnom svijetu. Zbog toga se pribjegava korištenju čovjeka kao učitelja i učenju od samog čovjeka (Perez-Dattari et al., 2020). Ovaj pristup djeluje obečavajuće te se možemo nadati da ćemo uskoro dobiti robote koji mogu da služe ljudima u svakodnevnom životu i prilagođavaju se trenutnim zadacima koji su pred njima.

\subsection{Samopodešavajuće mašinsko učenje}
Naš život je prepun složenih izbora koje svakodnevno donesimo kako bi maksizirali svoje interese. Pri donošenju optimalnog izbora, pojedinac sagledava trenutnu situaciju i rasuđuje koja opcija bi bila najbolja u tom trenutku. Ako krajnji cilj zavisi od velikog broja nezavisnih kombinacija, naše rasuđivanje će biti otežano pa na kraju i nemoguće. Kada farmaceutski istraživač dizajnira nove lijekove za borbu protiv određene bolesti, ne samo što mora odabrati kombinaciju koja je liječi već i onu kombinaciju koja je u osnovi sigurna za korisnika. Korištenjem AI dakako olakša naš izbor i omogućava nam uvid u nevjerovatne rezultate kako određena kombinacija može uticati na krajnji ishod. Ali i AI kao i većina drugih industrijskih postrojenja je podložna specifičnim izborima modeliranja koji na kraju mogu poboljšati ali također i pogoršati krajnji cilj.  U prvom redu tu naravno govorimo o umjetnim neuronskim mrežama čiji krajnji rezultat zavisi od našeg izbora arhitekture (tj. koliko smo nivoa učenja koristili, da li smo koristili jedan ili drugi način rješavanja jednačina, koliko smo dugo trenirali i slično) (Karamehmedović et al., 2019).

Samopodešavajuće mašinsko učenje (eng. AutoML) je specijalna grana AI gdje koristimo predikcione metode kako bismo efikasno pronašli optimalnu kombinaciju za numerički kompleksne sisteme kao što su mreže gdje direktna relacija između ulaznih dizajnerskih odluka i krajnjeg cilja je nepoznata (Shahriari et al., 2015), prikazano na slici 1. Predikcione metode nam omogučavaju da stvorimo pretpostavku kako se krajnji cilj ponaša u odnosu na ulazne dizajnerske odluke što za rezultate daje da možemo rasuđivati koja odluka bi potencijalna mogla biti optimalna za naš problem (tj. da li mreža treba imati 2 nivoa učenja ili više). Potencijalno optimalne odluke moramo evaluirati kroz kompleksni sistem kako bi se uvjerili u njenu kvalitetu. U slučaju da odluka ne zadovoljava kriterije, proces se ponavlja gdje prošle odluke koristimo kao određeni vid učenja koji će nas odvesti do optimalne odluke i optimalnog performansa.

Metode koje se koriste u AutoML  su u osnovici uopštene metode koje za cilj imaju pronaći optimalno rješenje bilo kojeg kompleksnog problema što efikasnije jer ovakvi i slični problemi su veoma računarski skupocijeni (Šehić et al., 2021). Povećavanjem broja dizajnerskih opcija smanjuje se efikasnost ovih metoda jer predikcione metode postaju eksponencijalno zahtjevnije. Određena moguća rješenja su primjena predikcionih metoda lokalno ili pronalazak niskodimenzionalnih projekcija (Šehić et al., 2021). Iako je zastupljenost AutoML u naučnim radovima i dalje veoma niska, za očekivati da će se to uskoro promijeniti jer AutoML metode postaju lahko pristupačne.

\subsection{Kompresija umjetnih neuronskih mreža}
Motivisani činjenicom da interakcije između moždanih ćelija nisu gusto povezane prilikom rješavanja konkretnih zadataka, istraživači u oblasti umjetne inteligencije uvidjeli su da slične principe mogu primijeniti i na sami razvoj i treniranje modela umjetnih neuronskih mreža.  Osim ove biološki inspirisane ideje, benefiti od kompresije ogromnih modela ogledaju se i u velikim uštedama električne energije potrebne za njihovo razvijanje i korištenje u industrijskim aplikacijama.

Veliki broj različitih metoda bi se mogao svrstati pod pojam kompresije umjetnih neuronskih mreža (Hoefler et al., 2021) i sve one imaju zajednički cilj: smanjenje računarskih zahtjeva ogromnih modela uz minimalno narušavanje njihovih performansi. Usko povezan sa trendom razvoja hardvera i dostupnosti sve veće količine podataka za mašinsko učenje, rapidni porast broja parametara u neuronskim mrežama nije kaskao. Tome pridonosi i činjenica da iz dana u dan pokušavamo trenirati neuronske mreže da rješavaju kompleksnije probleme, te se kao jedno od najlogičnijih rješenja utemeljilo povećanje ekspresivne moći modela kroz direktno povećavanje broja parametara. No međutim, imajući na umu da puko povećavanje broja parametara nije rješenje koje možemo pratiti daleko u budućnost, postavlja se pitanje da li je moguće pronaći modele koji bi sa mnogo manjim brojem parametara jednako dobro rješavali željene zadatke. Tokom vremena razvila su se dva pravca: 1) razvijanje malih modela sa kompresovanim parametrima od samog početka, i 2) identifikacija i kompresija redundantih parametara za određeni zadatak u već gotovim modelima. Ispostavilo se da je prvi pravac veoma zahtjevan zbog same činjenice da je ekspresivna moć modela umanjena od samog početka procesa učenja. Samim time to je nešto što ga čini primamljivijim istraživačima u ovoj oblasti jer zahtjeva nove ideje i pomak od trenutne paradigme. Drugi pravac, relativno jednostavniji, se uspostavio kao popularan metod medju praktičarima koji žele iskoristiti neuronske mreže za rješavanje konkretnih problema na uređajima s kojima se susrećemo u svakodnevnom životu (mobiteli, automobili, industrijski pogoni, pametni satovi, itd.). Ovaj metod se oslanja na već unaprijed razvijene modele koji se onda postepeno smanjuju i evaluiraju na konkretnom problemu dok se ne dostigne zadovoljavajući omjer cijene njihovog korištenja i kvalitete performanse. Jedan od primjera ovakvog pristupa kompresiji je predstavljen u radu (Kurtić et al., 2022), gdje su ogromni modeli za razumijevanje teksta kompresovani čak i do deset puta bez primjetnog gubitka performansi.  

I na kraju bitno je spomenuti nekoliko činjenica koje su jasni indikatori da se još uvijek nalazimo u ranoj fazi razvoja ove oblasti: 1) softverska podrška za treniranje kompresovanih modela je u ranom razvoju i još uvijek nemamo rješenje kako efikasno iskoristiti dostupni hardver, 2) hardver koji danas koristimo je razvijen s ciljem obrade velikih paketa podataka i samim time trenutno ne može najefikasnije iskoristiti činjenicu da su modeli kompresovani, 3) veliki broj industrijskih rješenja se još uvijek, nažalost, temelji na primitivnim metodama kompresije.

\subsection{Kombinovanje metaheurističkih algoritama i data mininga za poboljšanje procesa}
Razvoj online platformi i upotreba informacionih sistema značajno su povećali količinu prikupljenih podataka. Uz razvoj računske moći i jeftinu memoriju, omogućen je ubrzan razvoj oblasti koje se zasnivaju na upotrebi i analizi prikupljenih podataka, poput oblasti data mininga ili mašinskog učenja. Međutim, i dalje postoji niz praktičnih prolema koji se ne mogu riješiti navedenim metodama, poput teških optimizacijskih problema. Za njihovo rješavanje i garanciju optimuma, neophodno je pretražiti kompletan prostor. Istovremeno, pretraga prostora nije moguća u realnom vremenu, te se koriste metaheuristički algoritmi. Metaheuristički algoritmi se fokusiraju na pronalazak subotimalnog rješenja, rješenja koje se dobija u dovoljno kratkom vremenu, ali ne garantuje optimalnu vrijednost. Neki od najpoznatijih modernih metaheurističkih algoritama su Algoritam slijepog miša (BA), Algoritam svica (FA), Algoritam vatrometa (FWA), Algoritam krda slonova (EHO) (Osaba et al., 2021). Kombinacijom metaheurističkih pristupa, data mining tehnika i korištenjem prikupljenih podataka, moguće je značajno poboljšati procese i povećati mogućnost pronalaska optimuma.

U radu (Delalić et al., 2020), opisan je problem planiranja događaja korištenjem podataka prikupljenih na društvenim mrežama. U radovima se koriste agregirane geografske lokacije i broj pratilaca po svakom gradu, te se dobijeni podaci koriste za određivanje lokacija održavanja događaja.

Kreiran je koncept pametnog sistema upravljanja skladištem (WMS). Predloženo je poboljšanje koncepta pametnog skladišta korištenjem algoritma slijepog miša za spajanje narudžbi i njihovo grupno prikupljanje. 

Sistem upravljanja skladištem proširen je planiranjem transportnih ruta, pri čemu se ranije prikupljeni GPS podaci koriste za predviđanje vremena istovara robe, kao i za niz drugih poboljšanja, ali i primjenom metaheurističkih algoritama za klastering kupaca i rješavanje problema VRP-a u dvije faze (Žunić et al., 2022).

Kompletan sistem planiranja transportnih ruta i upravljanja skladištem sačinjava značajan dio SCM-a (engl. Supply Chain Management). Opisani koncept pametnog SCM-a implementiran je u više distribucijskih kompanija u Bosni i Hercegovini i regiji i detaljno je opisan u (Žunić et al., 2021).

\subsection{Učenje sa podsticajem}
Kao što je navedeno u poglavlju “Šta je AI”, učenje s podsticajem je grana umjetne inteligencije koja pretpostavlja interakciju između AI (tzv. agenta) i okoline u kojoj AI djeluje, za razliku od drudge dvije grane AI koje se bave problemima predikcije. Računarske igre su historijski najčešće korištene okoline za razvoj i testiranje algoritama učenja sa podsticajem. Jedan od prvih takvih algoritama koji igra na nivou ljudskog igrača, tzv. TD-Gammon (Tesauro, 1994), nastao je 1992. za igru dama. U zadnjih nekoliko godina, vidjeli smo novije algoritme koji koriste učenje sa podsticajem i premašuju ljudske sposobnosti. Tako je 2014. godine DeepMind kreirao algoritam Deep Q-Learning (Mnih et al., 2013) koji ostvaruje impresivne rezultate u domenu Atari igrica, te AlphaGo algoritam 2015. koji premašuje najbolje svjetske igrače u možda najkomplikovanijoj društvenoj igri, kineskom Go (Silver et al., 2016).

Korištenjem metoda učenja sa supervizijom, DeepMind je naknadno, 2020. godine objavio generalizaciju AlphaGo algoritma pod imenom MuZero (Schrittwieser et al., 2020), koji pomoću učenja modela okoline uspješno premašuje ljudske sposobnosti u Atari igrama, te igri Go i šahu. Modeliranje okoline se bazira na učenju sa supervizijom s ciljem predviđanja stanja okoline u budućnosti - npr predikcija broja bodova u igri ukoliko igrač odigra određeni niz poteza. Sami po sebi takvi modeli već doprinose boljem razumijevanju okoline i samim tim boljim rezultatima u nizu igara (Gregor et al., 2019), ali pored toga, modeli okoline se mogu koristiti za planiranje, što algoritmi poput MuZero i demonstriraju.

Mnogi stručnjaci su mišljenja da je učenje sa podsticajem put ka generalnoj umjetnoj inteligenciji (Silver et al., 2021). Velike svjetske firme kao i nekolicina država vode žestoku borbu na putu ka generalnoj umjetnoj inteligencije - nešto što mnogi svjetski stručnjaci porede sa trkom na mjesec. Priznati naučnici hipotetiziraju da je modeliranje okoline neophodno za razvoj zaista generalne umjetne inteligencije, ali da modelima kojima trenutno baratamo nedostaje razumijevanje uzročno-posljedičnih odnosa (Pearl and Mackenzie, 2018), koji su prema njima u srži generalne inteligencije.

\subsection{Efikasne hardverske platforme za AI sisteme}
Tri ključna faktora za razvoj AI-a su algoritamska rješenja, dostupnost velike količine podataka i moćni računarski sistemi na kojima je moguće izvršenje AI algoritama. Od ova tri faktora, razvoj namjenskih računarskih sistema omogućio je praktičnu primjenu AI algoritama u širokom spektru aplikacija, uključujući transport, medicinu, sigurnost, ekonomiju, internet usluge. Iako su grafički procesori (eng. Graphical Processing Unit, GPU) doveli do značajnog povećanja efikasnosti izvršenja AI algoritama, reducirajući i vrijeme i utrošak energije, danas AI opet ima hardverski problem (“Does AI have a hardware problem?,” 2018). Historijski (do 2012.), potreba za racunarskim kapacitetima se udvostručavala svake dvije godine, međutim danas je ovaj trend dosta brži i udvostručava se svaka 2 mjeseca. Ovaj trend je nemoguće pratiti tradicionalnim rješenjima (npr. Moore-ov trend smanjivanja elektroničkih tranzistora dovodi do udvostručavanja računarske moći svake dvije godine). Inovacije u računarskoj arhitekturi i dizajnu novih čipova (eng. chip) dovele su do značajnijeg povećanja efikasnosti – preko 300 puta u periodu od 2012. do 2020. godine - ali ovo i dalje nije dovoljno da bi se pratio trend udvostručavanja svaka 2 mjeseca. Direktni rezultat ovoga je da se cijena treninga zahtjevnih AI algoritama povećala od par dolara u 2012. godini do preko 10 miliona dolara u 2020. godini (Mehonic and Kenyon, 2022). Ovaj trend nije održiv. Pored toga, trening kompleksnih AI algoritama na današnjim računarskim sistemima (npr. globalnim internet serverima) zahtijeva ogromne energetske resurse i ima karbonski otisak ekvivalentan emisiji pet automobila (Strubell et al., 2019).

Sve ovo je dovelo do velikog interesovanja u razvoj alternativnih fundamentalnih elektroničkih komponenata za efikasnije AI sisteme. Jedna od najpopularnijih oblasti razvoja su takozvani neuromorfološki sistemi, čija funkcionalnost je inspirisana biološkim sistemima poput ljudskog mozga – sistem mnogostruko energetski efikasniji od danasnjih najmočnijih računara. Neuromorfološki inženjering podrazumijeva razvoj novih elektroničkih materijala i uređaja (npr. memristorskih uređaja), novih kompjuterskih arhitektura, te efikasnijih AI algoritama. Neuromorfološki sistemi se značajno razlikuju od današnjih digitalnih računarskih sistema: ovi sistemi mogu biti asinhroni, analogni, vršiti obradu podataka direktno u memoriji, te nisu bazirani na logičkim kolima. Ovi sistemi su u ranoj fazi razvoja, ali prvi prototipi pokazuju obećavajuću efikasnost, više od 100 puta veću nego najefikasniji računari današnjice. Razvoj ovakvih sistema nije ograničen samo na akademske istrazivačke grupe, vec se razvijaju i u industriji. Neki od primjera industrijskih sistema su IBM-ov True North neuromorfološki čip i INTEL-ov LOIHI čip. Neuromorfološki čipovi i sistemi mogu dovesti do značajne prekretnice u razvoju AI sistema sa veoma niskom potrošnjom energije (npr. u mobilnim uređajima, pametnim senzorima, mikrouređajima).

\begin{figure}[t]
  \centering
  \includegraphics[trim=0 0 0 0,clip, width=1.0\linewidth]{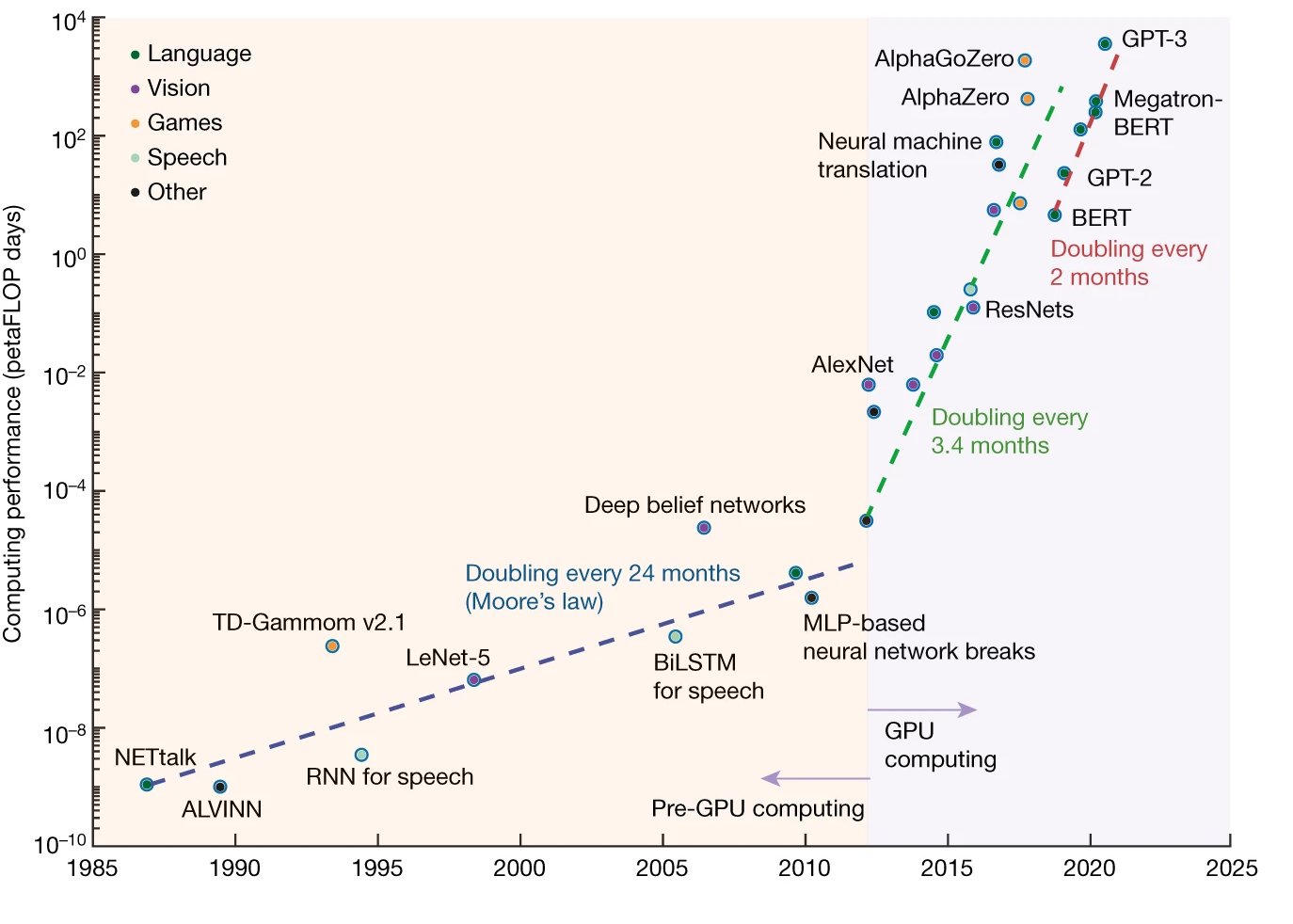}
  \caption{Zahtjevi za računarskom snagom (Mehonic and Kenyon, 2022).}
  \label{fig:adnan}
\end{figure}

\section{Aktualni trendovi}
Promatranje trendova omogućava bolje razumijevanje situacije te samo pozicioniranje, što je posebno izazovno u vremenu brzih promjena i mnoštvu informacija. Stoga u ovom poglavlju dajemo kratki uvid u aktualne trendove koji se mogu primjetiti vezano za AI. To uključuje same trendove vezane za AI metode, alate i infrastukturu; paralelne trendove koji se oslanjaju na AI te društvene trendove vezane za AI.

Kao osnovni, veoma je uočljivo nekoliko trendova razvoja AI metoda. U oblasti ML, kao jednoj od nosioca razvoja AI današnjice, evidentno je da su ML modeli sve veći (npr. sadrže i preko 500 milijardi parametara (Chowdhery et al., 2022)) te se treniraju na sve većim skupovima podataka (npr. ImageNet sadrži 14 milijuna ručno označenih slika (Deng et al., 2009)). Sa druge strane, da bi se riješio problem transparentnosti i objašnjivosti ML modela, pristupa se razvijanju metoda inspirisanih simboličkom AI, tj. tkz. neuro-simboličke metode  (engl. neuro-symbolic AI).  Trend razvoja AI metoda prate i trendovi razvoja alata i infrastrukture za razvoj i korištenje metoda. Tako se može primjetiti da je sve više alata dostupno u vidu otvorenog koda (engl. Open source) kao što su TensorFlow, Pytorch, i drugi. Ovi alati uveliko olakšavaju razvoj i korištenje kao i samo poređenje različitih AI metoda. Za razvoj AI hardvera može se reći da je davno prevazišao prvobitni razvoj GPU koje su korištene u računarskim igrama, tako da se za treniranje ML modela koriste namjenski računarski sistemi za TPU čipovima (engl. Tensor Processing Unit). Trend rasta računarkih zahtjeva se rapidno ubrzava, tako da se trenutno računarski zahtjevi za treniranje savremenih modela udvostručuju svako dva mjeseca. Kako je sam napredak doveo je do evidentih uvečanih zahtjeva za energijom, postaje dominantna teznja da se razvijaju energetski-efikasne hardverske strukture (Mehonic and Kenyon, 2022) (npr. bazirane na analognim tehnologijama). 

Sam trend razvoja AI metoda i dostupne infrastrukture uzročno prati trend rješavanja sve kompleksnijih problema. Tako su uspješno riješeni problemi koji su nekada smatrani nerješivima za AI. Tako je AI ostvarila performance iznad eksperta čovjeka u igri šaha, prepoznavanju objekata na slikama ImageNet (Krizhevsky et al., 2012), igra Go (Silver et al., 2016), savijanju proteina (Jumper et al., 2021), itd. AI sa velikim jezičkim modelima (engl. Large Language Models - LLM) je sposobna da u ograničenim okolnostima vodi smislene razgovore sa covjekom i generiše tekst na zadanu temu. Također, AI se koristi u ograničenim funkcijama za automatiziranu vožnju, prvenstveno za percepciju mašinskom vizijom. 

Osim samih trendova vezanih za razvoj metoda, alata i infrastructure za AI, trenutno se mogu uočiti paralelni trendovi koji uveliko utiču na razvoj AI te se ujedno oslanjaju na sam napredak AI. Ovo su trendovi razvoja u ostalim tehnološkim poljima koji se isprepliću sa AI na naćin da koriste određene AI metode u svojim primjenama, ali i pružaju podatke ili naćine da AI utiće na različite aspekte naših života. U segmentu zabave i računarskih igara, zadnjih godina evidentatan je trend razvoja Augmentirane i Virtuelne Realnosti. Ova tehnologija koristi AI u različitim segmentima (npr. pračenje zjenica ili kretanja glave, rendering virtualnog svijeta itd.) ali i omogučava poboljšan protok informacija od računara prema korisniku. Sa druge strane, uočljiv je trend razvoja interfejs (eng. interface) tehnolgija za olakšano upravljanje digitalnim sistemima i protok infomacija od korisnika prema računaru. Primjeri ovakvih sučelja uključuju govorno sučelje (npr. Alexa) ili mozak-računar interfejs (eng. brain-computer interface - BCI). Digitalizacija i otvaranje pristupa različitim sektorima, omogućava pristup podacima za razvoj preciznijih modela i primjenu AI u realnim sekotirma na način da sistemi budu adaptabilini za posebne situacije i potrebe ili da se optimizuju prema nekom kriteriju. Tako je uočljiv trend razvoja tkzv. pametnih gradova (engl. Smart city), različitih pametnih uređaja (IoT uređaji), Industrije 4.0 i 5.0, Inteligentni transportni sistem (ITS). Blockchain tehnologija i sistemi bazirani na njima  (npr. Bitcoin) otvaraju pristup svim korisnicima finansijskom sektoru i čini javno dostupnim podatke o transakcijama te samim time i različite aplikacije AI. Te sami uređaji koje korisnici koriste kao što su pametni satovi (engl. wearables) pružaju podatke o pojedinim korisnicima korisne za personaliziranje usluga za samog korisnika.  Pored toga, napredak bežične komunikacije sa 4G, 5G i 6G omogučava pouzdaniji i brži pristup podacima u realnom vremenu.

Pored tehnoloških trendova veoma je važno uočiti i društvene trendove vezano za AI. Da bi u potpunosti iskoristili potencijale koje nosi AI i izbjegli potenicijane negativne posljedice, gotovo sve zemlje su razvile ciljane strategije te pokrenule sveobuhvatne inicijative sa dugogodišnjim akcijama. Kroz strategije, zemlje pokušavaju da iskoriste svoje kompetitivne prednosti te da se bolje pozicioniraju i iskoriste prilike koje sama AI nosi. Zbog posebne važnosti u sljedećem poglavlju smo detaljnije predstavili neke od nacionalnih AI strategija.

\section{Nacionalne AI strategije}
Kroz ovaj rad predstavljamo pregled strategija nekih od zemalja iz našeg iskustva. To uključuje zemlje Evrope (Austrija, Švedska, Danska, UK i Nizozemska), Australija, Sjedinjene Američke Države te zemlje susjedstva uključujući i  ograničene aktivnosti u Bosni i Hercegovini. Pregled AI politika je dostupan na OECD.AI portalu (EC/OECD, 2021).

\subsection{Austrija}
Austrijska stratagija je definisana u dokumentu “Artificial Intelligence Mission Austria 2030 (AIM AT 2030)” (Federal Ministry Republic of Austria for Digital and Economic Affairs (BMDW), 2021). U dokumentu je jasno naglašeno da misija/dokument može i treba da se kontinuirano modifikuje i evoluira vremenom prateći trendove naučnih dostignuća. Fokus je dat na tri ideje:
\begin{itemize}
    \item upotreba AI-baziranih rješenja za unaprijeđenje svakodnevnog života (npr. unaprijeđenje administrativnih sistema),
    \item pozicioniranje Austrije kao centra za inovacije i istraživanje u pogledu AI (podržavanje i univerziteta i start-up sfere),
    \item iskorištavanje AI-based rješenja za povećanje kompetitivnosti “domaćih” tehnoloških proizvoda.
\end{itemize}

U samom pisanju strategije učestvovao je veliki broj stručnjaka iz raznih oblasti (preko 160 stručnjaka iz oblasti tehnologije, ekonomije, prirodnih i društvenih nauka), veliki broj ministarstava, eksperti iz struke, i stakeholderi vodećih kompanija. Ponuđeno je nekoliko primjera podrške vlade naučnim institucijama: podrška otvaranju IST Park-a, startup hub-a pri istraživačkom institutu IST Austria, s ciljem monetizovanja naučnih dostignuća. Posebno zanimljiva je podrška vlade otvaranju research hub-ova velikih kompanija kao što su Amazon, Facebook, Snapchat. Vezano za infrastrukturu, obezbijeđeno je kontinuirano poboljšanje Vienna Scientific Cluster-a, najmoćnijeg super-računara u Austriji. Osnovan je kao zajednički projekat nekoliko Austrijskih univerziteta sa velikom finansijskom podrškom Austrijskog ministarstva za edukaciju, nauku i istraživanje. 

\subsection{Švedska}
Švedska Vlada je u maju 2018. godine usvojila nacionalnu strategiju po pitanju AI (Government Offices of Sweden, 2018) u cilju da definiše opšte pravce na državnom nivou a koji bi služili kao osnovica za definisanje politika i prioriteta u Švedskoj. Nacionalni plan se fokusirao na 4 elementa obrazovanje, istraživanje, inovacije i primjena, i infrastuktura. Prije usvajanja samog nacionalnog plana, švedska državna agencija za inovacije Vinnova je prezentirala detaljan izvještaj (Vinnova, 2018) o mogućnostima Švedske na polju AI kao i samim izazovima.  Samo u u 2020. godini, Vinnova je podržala projekte AI-a sa 675 miliona SEK (okvirno 130.6 miliona BAM). Ukupne investicije prelaze 1.35 miliarde SEK (okvirno 261.2 miliona BAM) dok nacionalni proračun planira do 2024 godine investirati najmanje 550 miliona SEK (okvirno 106.4 miliona BAM) za istraživanje i inovacije u AI i njegovu primjenu i uticaj na društvo. Pored integrisanja novih studijskih ciklusa na švedskim univerzitetima koji se fokusiraju na AI, cilj je ponuditi također i programe prekvalifikacija i dodatnog usavršavanja. Također, u suradnji sa industrijom, švedska Vlada želi stvoriti bolje uslove da industrija uvidi cjelokupne benefite primjene AI-a u proizvodnji. Dodatno, Vinnova je podržala čak 256 inovativna projekta kao i određene startup ideje. Strategija jasno definiše korake za promjenu i poboljšavanje zakonskih rješenja kako bi se olakšala primjena i rad na AI-u u Švedskoj kao i prijedloge za bolje umrežavanje institucija u vezi AI-a. 

\subsection{Danska}
Danska Vlada je svoju strategiju (Ministry of Finance and Ministry of Industry, Business and and Financial Affairs of Denmark, 2019) usvojila u martu 2019. godine sa četiri glavna cilja a to su razvoj etičkih osnova za AI, podržati istraživanja u AI kao i privatni sektor u njegovoj primjeni, i implementiranje AI kao sastavnog dijela javnog sektora što bi za rezultat trebalo dati kvalitetnije usluge za građane. Strategija obuhvata 24 inicijative za koje je danska vlada obezbjedila preko 18 miliona BAM u periodu 2019-2027. Prioritet ovih inicijativa je većinom ka zdravstvu, energiji, poljoprivredi i transportu. Nadalje, cilj je uskladiti obrazovni sistem tako da ukljući direktno AI u nastavni proces preko upoznavanja učenika sa upotrebom i korištenjem AI-a u svakodnevnici do jasno definisanih studijskih ciklusa. Dodatni fokus strategije je također stavljen na umrežavanje između privatnog i javnog sektora kako bi javni sektor iskoristio u potpunosti resurse i iskustva privatnog sektora za dobrobit samih građana. Kako je jedan od glavnih ciljeva razvoj etičkog okvira za AI-a, strategija predlaže šest principa kako bi se poboljšao nivo povjerenja u AI. Potrebno je osigurati da građani mogu donositi nezavisne odluke i da nema kršenja osnovnih ljudskih prava. Dodatno, primjena AI mora biti otvorena, transparentna i etički odgovorna. Kako je jedan od bitnijih preduvjeta za razvoj AI-a posjedovanje kredibilinih podataka, danska vlada se obvezuje da olakša pristup i stavi na raspolaganje vlastite podatke iz oblasti zdravstva, javnog sektora i slično građanima, privatnom sektoru i istraživačima.

\subsection{Nizozemska}           
U Oktobru 2019 godine nizozemska vlada je objavila strategijski akcioni plan za umjetnu inteligenciju (eng. Strategic action plan for artificial intelligence) sa ciljem prezentiranja spektra politika koje bi doprinijele kompetitivnosti Nizozemske u svijetu AI-a (The Netherlands, Ministry of Economic Affairs and Climate Change, 2019). Vizija nizozemske strategije se oslanja na tri stuba:
\begin{itemize}
    \item kapitalizacija socijalnih i ekonomskih prilika,
    \item kreiranje odgovaranjućih uslova,
    \item pojaćavanje osnova.
\end{itemize}

Strategija sadrži proširenu listu inicijativa koje imaju za cilj da potaknu AI u ekonomiji kroz politike povezane za obrazovanje, istraživanje, razvoj i inovacije, te povezivanje (eng. Networking), regulacije i infrastrukturu.

Godišnji buđet za AI inovacije i razvoj je bio otprilike 45 milliona eura godišnje (prije 2019 godine).  U 2019. godini buđet je bio 64 miliona eura, u 2020 dodatnih 23.5 miliona eura za privatno-javno partnerstvo Netherlands AI Coalition (NL AI Coalition, 2022). NL AIC okuplja preko 400 učesnika u nacionlanom AI programu. U Aprilu 2021 je objavljen investicijski program AiNed (AiNed Foundation, 2021) koji omogucava NL AIC da se buđet poveća do maksimalnih 276 miliona eura (samo u prvoj fazi AiNed projekta) u narednim godinama, što pokazuje opredjeljenost nizozemske vlade za unapređenje ulaganja u AI. AiNed ima za cilj da ubrza razvoj i upotrebu  AI tehnologija. Fokusiran je na projekte velike skale za: 1) ubrzavanje inovativnih aplikacija AI, 2) ojačavanje baze znanja fundamentalnog i primjenjenog istraživanja, 3) povećanje kapaciteta za AI obrazovanje i trening, 4) razvoj AI fokusirane na čovjeka (engl. Human-Centred) sa etičkim i pravnimokvirima i 5) činjenjem podataka dostupnih za AI.
Pregled nizozemske AI politike je dostupan na OECD.AI portalu (EC/OECD, 2021).

\subsection{Ujedinjeno Kraljevstvo}
U oktobru 2017. godine, vlada Velike Britanije angažovala je tim stručnjake iz oblasti umjetne inteligencije da sastave izvještaj o trenutnom stanju i budućem potencijalu ove rastuće tehnologije (Hall and Pesenti, 2017). Kao rezultat tog istraživanja, vlada Velike Britanije je u aprilu 2018. objavila (i finalizirala u maju 2019.) nacionalnu strategiju za rast i razvoj umjetne inteligencije, te oformila ured posvećen sprovođenju iste.
Cilj strategije je prvenstveno etablirati Veliku Britaniju kao vodeću naciju u oblasti umjetne inteligencije, te pripremiti ekonomiju i društvo u cjelini za promjene koje dolaze kao rezultat razvoja ove i sličnih tehnologija. Fokus navedene strategije je podijeljen u pet ključnih oblasti:
\begin{itemize}
    \item Inovacije,
    \item Ljudski kapital,
    \item Infrastruktura,
    \item Ekonomski razvoj,
    \item Prosperitetna zajednica.
\end{itemize}
Izvještaj navodi da je za stimuliranje inovacija planirano dugoroćno povećanje budžeta za istraživački rad u javnom sektoru na 3\%, te uspostavljanje investicijskog fonda od 2.5 milijarde namijenjenog za rast i razvoj privatnog sektora. Pored toga, strategijom se uvode dodatne poreske olakšice firmama koje se bave istraživačkim radom. Za unaprjeđenje ljudskog kapitala, strategijom je planirano povećano ulaganje u Alan Turing institut za umjetnu inteligenciju, te kreiranje preko 1000 novih doktorskih i preko 2500 magistarskih pozicija u narednih pet godina. Uz to, alociran je budžet za mnoštvo novih stipendija i istraživačkih grantova u oblasti umjetne inteligencije, te nove prilike za prekvalificiranje u STEM oblasti. Strategija navodi i detaljne planove za razvoj infrastrukture, od povećanja generalne digitalizacije društva kroz pristupačnije i brže internet konekcije (fiber internet i 5G mobilne mreže), do povećanja komputacijskih kapaciteta i formiranja novih superkompjutera za treniranje modela za umjetnu inteligenciju.

Strategija posvećuje posebnu pažnju krucijalnoj ulozi kvalitetnih podataka, te navodi mjere poboljšanja infrastrukture za pohranjivanje i generalnu pristupačnosti istih. Strategija ide i korak dalje i najavljuje formiranje instituta za javnu pristupačnost podataka s ciljem da ne samo obezbijedi pristup korisnim podacima i metodama upotrebe istih, već i uveđenja standarda za fer upotrebu podataka koja će koristiti društvu u cjelini.

U totalu, procjenjuje se da strategija u narednom periodu ima za cilj povećanje ulaganja u oblast umjetne inteligencije kroz 0.95 milijardi funti direktnog ulaganja, te 1.7 milijardi funti kroz ulaganja u povezane oblasti i generalnu infrastrukturu. Ova opširna strategija o umjetnoj inteligencije predstavlja jednu od prvih svoje vrste i potvrđuje bitnost podataka i pametne upotrebe istih za razvoj društva i ekonomije u cjelosti.

\subsection{Australija}
U Novembru 2019 Australijska vlada je objavila izvještaj koji je predstavio nacionalnu AI strategiju sa ciljem povećanje produktivnosti industrije, otvaranje novih radnih mjesta i ekonomskog rasta te poboljšanja kvalitete života sadašnjih i budučih generacija (Hajkowicz et al., 2019). U izvještaju su predstvljena tri strateška pristupa u realizaciji postavljenih ciljeva: i) AI specijalizacija za upravljanje prirodnim resursima, unapređenje zdravstvenog sistema i  unaprjeđenje gradova i infrastrukture koja ima za cilj povećenje produktivnosti, značajno smanjenje troškova te poboljšanje uslova, sigurnosti i kvalitete života.

ii) Istraživanje vođeno misijom koje ima za cilj da identifikuje i riješi nacionalne probleme i izazove, pri čemu bi misija mogla da nadopunjuje specijalizaciju. Ovaj pristup bi mogao biti efiksan način razvoja mogućnosti AI dok u isto vrijeme bi onogućio rješavanje problema od nacionalnog interesa i važnosti. iii) Ekosistem koji obuhvata postojeći privatni i javni sektor, resurse i/ili pojedince vezane za poslovne ciljeve i AI. U junu 2021 predstavljen je AI akcioni plan (Department of Industry, 2021). Direktne mjere Akcionog plana uključuju investiciju australske vlade od 124,1 miliona dolara za jačanje australskog liderstva u razvoju i primjeni odgovorne AI u periodu od 5 godina. Plan se fokusira na četiri ciljna područja: i) unaprijediti razvoj i primjenu AI radi otvaranja radnih mjesta i povećanja produktivnosti, ii) rasti i privući talente i stručnjake na globalnom nivou, iii) iskoristiti vlastite vodeće svjetske AI sposobnosti za rješavanje nacionalnih izazova, iv) osigurati odgovornu i inkluzivne AI tehnologiju. 

\subsection{Sjedinjene Američke Države}
Kao vodeća država svijeta u pogledu tehnološkog i ekonomskog razvoja, Sjedinjene Američke Države (SAD) su također prepoznale izuzetan potencijal umjetne inteligencije. Koliki se značaj pridaje ovoj oblasti pokazuje i sama činjenica da je od prošle godine sa radom počela posebna internet stranica američke vlade posvećena ovoj oblasti (https://www.ai.gov/). U svom godišnjem obraćanju naciji 5. februara 2019. godine tadašnji Predsjednik Donald Trump je istakao značaj očuvanja američke vodeće pozicije u razvoju i upotrebi tehnologija 21. stoljeća, uključujući i umjetnu inteligenciju.  Nedugo nakon toga, 11. februara 2019. godine, Predsjednik Trump izdaje Predsjednički ukaz broj 13859 pod nazivom “Očuvanje američkog liderstva u umjetnoj inteligenciji” (eng. Maintaining American Leadership in Artificial Intelligence) (“Maintaining American Leadership in Artificial Intelligence,” 2019). Ovim ukazom je ustanovljena Američka inicijativa za umjetnu inteligenciju (eng. the American Artificial Intelligence Initiative), a neke od njenih glavnih direktiva su povećanje ulaganja u AI istraživanja, uspostavljanja tehničkih standarda, povećanje dostupnosti računarskih resursa i podataka, obuka odgovarajućeg kadra, kao i uspostavljanje međunarodne saradnje. Iste godine objavljen je i ažurirani Nacionalni strateški plan za istraživanja i razvoj u oblasti umjetne inteligencije (Science and Intelligence, 2019), a ranije ove godine je objavljen poziv za dostavljanje prijedloga kako bi se pristupilo njegovom ažuriranju (“National Artificial Intelligence R\&D Strategic Plan -- OSTP invites input for update (by 3/4) - EconSpark,” 2022). 

Bitno je istaći i da je u januaru 2021. godine sa radom počeo i Ured za nacionalnu inicijativu iz oblasti umjetne inteligencije (National Artificial Intelligence Initiative Office) koji djeluje pri Bijeloj kući i služi kao centralno tijelo za koordinaciju i kolaboraciju između državnih agencija, akademskih ustanova i privatnog sektora o pitanjima vezanim za umjetnu inteligenciju (“Artificial Intelligence for the American People,” 2021).

Pored primjena koje za cilj imaju poboljšanje svakodnevnog života građana i olakšavanje poslovanja kompanija, američka ulaganja u umjetnu inteligenciju su također podstaknuta i velikim ulaganjima u istu od strane njenih glavnih rivala, Kine i Rusije. U ovom kontekstu je značajno spomenuti i ulaganja od strane američke vojske, koja već duže vrijeme raspolaže naprednim tehnologijama poput dronova i bespilotnih letjelica, a njen budžet samo u 2021. godini je iznosio 768 milijardi američkih dolara. Za 2022. fiskalnu godinu, Ministarstvo odbrane je od Kongresa tražilo oko 14.7 milijardi dolara za projekte vezane za nauku i tehnologiju, od čega 874 miliona dolara za projekte iz oblasti umjetne inteligencije. (“Pentagon to spend \$874 million on artificial intelligence (AI) and machine learning technologies next year | Military Aerospace,” 2021). Prije par godina je otvoren i Zajednički centar za umjetnu inteligenciju (Joint Artificial Intelligence Center) koji za cilj ima istraživanja direktne primjene umjetne inteligencije na bojnom polju (“About the JAIC - JAIC,” 2019). Sve ovo sugeriše na veliki značaj koji će ove tehnologije imati u budućnosti kako u očuvanju nacionalne sigurnosti SAD-a, tako i u ostvarivanju njenih strateških geopolitičkih interesa širom svijeta.

\subsection{Bosna i Hercegovina i susjedne države}
U Bosni i Hercegovini i okolnim državama se stanje u oblasti AI postepeno mijenja i prilagođava uslovima EU. Povećanjem globalne popularnosti i broja stručnjaka iz oblasti AI, uočeno je povećanje u broju projekata i uočen je povećan interes na tržištu za razvoj AI.

U Hrvatskoj je pokrenuta inicijativa za kreiranje “Hrvatske nacionalne strategije za razvoj umjetne inteligencije”. U međuvremenu, u dokumentu koji definiše “Nacionalnu razvojnu strategiju Republike Hrvatske do 2030. godine” značajna uloga se posvećuje rastućoj ulozi AI. Razvojna strategija oblast AI vidi kao jedan od glavnih pokretača razvoja unutar države (Hrvatski sabor, 2021)\footnote{https://www.croai.org/regulation?lang=hr}.

U Srbiji je 2020. godine definisana strategija za razvoj AI do 2025. godine. Strategija se uklapa u Evropsku inicijativu za AI. U okviru strategije, najavljeno je osnivanje instituta, ali i niz primjena u industriji, obrazovanju i razvoju ekonomije (srbija.gov.rs, 2020).
U Bosni i Hercegovini je Parlament FBiH u dokumentu sa strategijom razvoja od 2021. do 2027. godine izdvojio značaj razvoja AI i prilagođavanja okvira strategiji razvoja u EU. Navedeno je da će osim uočavanja oblasti sa većim potencijalom primjene AI, veliku ulogu imati i osnivanje instituta za AI i uspostavljanje platforme za AI na nivou FBiH („Službene novine Federacije BiH“, broj 40/22, 2021).

\section{Naša vizija i realistične perspektive za Bosnu i Hercegovinu u doba AI}
Nakon detaljne analize stanja razvoja AI u svijetu kao i u Bosni i Hercegovini, kao glavni doprinos ovoga rada, u ovom poglavlju sintetiziramo našu viziju Bosne i Hercegovine u doba AI. Viziju pišemo imajući u vidu ograničene ekonomske mogućnosti i nedostatak odlučne poličke volje sa jedne stane, ali i urgentnost situacije te potrebu za djelovanjem da bi se izbjegao rizik kaskanja te gubljenje nivoa ostvarenog tehnološkog napredka u odnosu na ostatak svijeta. Stoga se fokusiramo na realistične perspektive koje efikasno koriste postojeće resurse i kompetitivne prednosti Bosne i Hercegovine te pružaju najbolji povrat investicije. 

\subsection{Vizija}
U duhu kolokvijalne teze “AI je električna energiju 21. stoljeća” (Ng, 2018),
Bosnu i Hercegovinu vidimo kao zemlju koja se kreće u korak sa Evropskim i svjetskim zemljama.  Održavajući tradiciju pionirskih poduhvata, nakon pionirske upotrebe električnih tramvaja u 19-om stoljeću, u 21-om stoljeću BiH aktivno doprinosti razvoju AI i odlučno primjenjuje AI tehnologije za dobrobit njenih građana.

\subsection{Prioritetne oblasti djelovanja}
Na osnovu prezentiranih analiza i posljedično formiranom stručnom mišljenju, prioritetne oblasti djelovanja za ostvarenje predstavljene vizije i pozicioniranje BiH na svjetskoj AI sceni se mogu definisati kao: 
\begin{itemize}
    \item Jasna strategija za dugogodišnji razvoj, 
    \item Pristupačna edukacija, 
    \item Infrastruktura za istraživanje, razvoj i komercijalizaciju AI, 
    \item Uslovi za privlačenje i zadržavanje kompetitivnog ljudskog kapitala,
    \item Uslovi i prilike za primjene AI u industriji. 
\end{itemize}

\begin{figure}[b]
  \centering
  \includegraphics[trim=0 0 0 0,clip, width=0.8\linewidth]{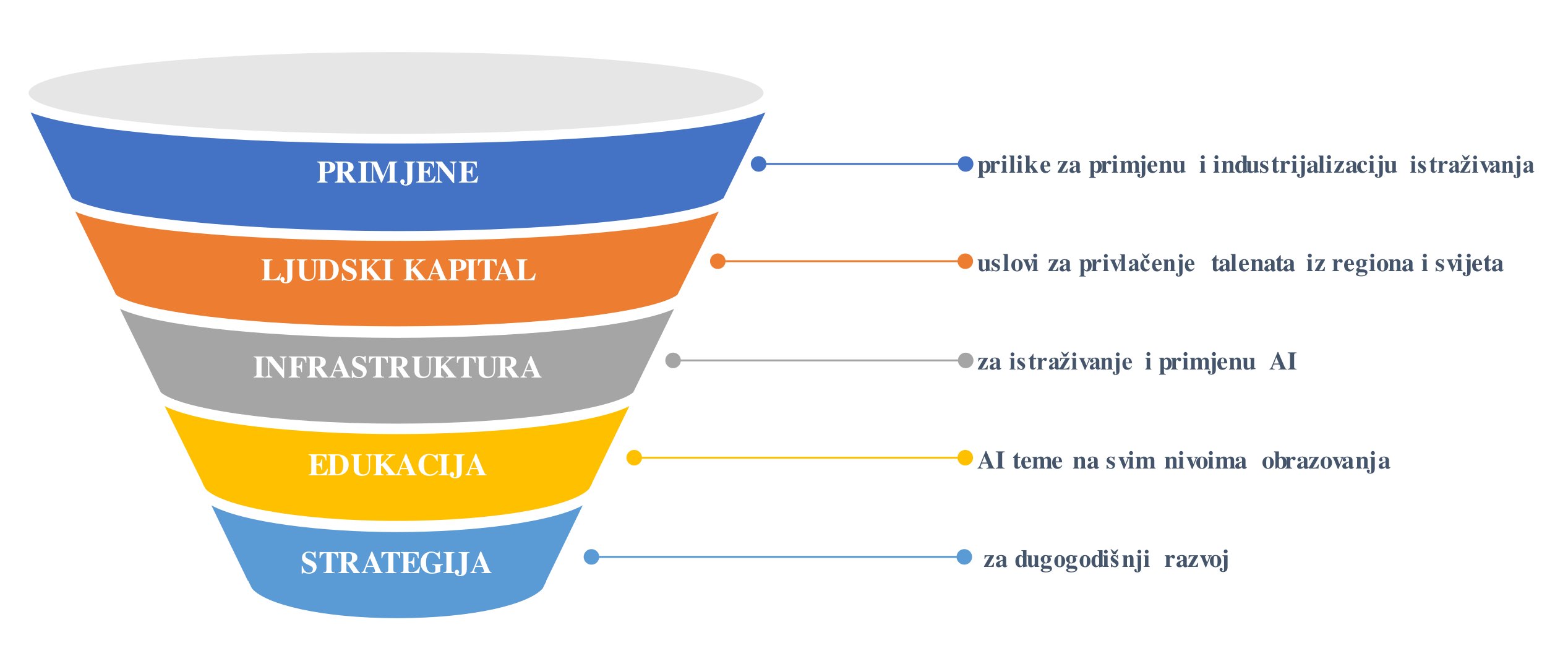}
  \caption{Prioritetne oblasti djelovanja na ostvarenju AI vizije.}
  \label{fig:vision}
\end{figure}

Prethodno navedene oblasti su vizuelno prikazane na slici \ref{fig:vision}.

\subsubsection{Jasna strategija za dugogodišnji razvoj}
Prvenstveno, vjerujemo da je potrebno razviti jasnu strategiju po uzoru na predstavljene primjere drugih zemalja koja bi omogućavala dugogodišnji razvoj AI u Bosni i Hercegovini. Sam proces izrade strategije, kao dio od krucijalnog značaja, treba biti inkluzivan. Tokom prikupljanja informacija i mišljenja i same izrade strategije potrebno je uključiti sve relevantne aktore, uključujući akademsku zajednicu, javnu upravu, predstavnike industrije i civilnog društva kao i neovisne strane eksperte. Posebnu pažnju bi trebalo posvetiti brendiranju strategije i njene vidljivosti sa ciljem kreiranja pozitivnog ugleda u Bosni i Hercegovini kao i inostranstvu. Cilj strategije treba biti animiranje svih postojećih resursa i konsolidacija trenutnih napora te ostvarenje samoodrživosti inicijative. Uz strategiju potrebno je definisati mehanizme i odgovorne za mapiranje čitavog bh AI ekosistema, praćenje provedbe i eventualne prilagodbe strategije.

\subsubsection{Pristupačna edukacija}
Da bi se adekvatno odgovorilo dinamičnim promjenama uzrokovanim rapidnim razvojem AI, iskoristile prilike i izbjegle eventualne opasnosti, potrebno je da se bh društvo adekvatno edukuje. Edukacije iz oblasti AI treba da budu pristupačne svim građanima sa različitim nivoima obrazovanja i profesijama, a to uključuje formalni obrazovni sistem (svi nivoi), neformalno obrazovanje i cjeloživotno učenje kao i šira javnost. Pristupačna edukacija je posebno bitna da bi se razvila fleksibilnost za budući dinamični razvoj AI oblasti, ali i održala kompetitivna prednost. Prvenstveno uzimajući u obzir da AI omogućava automatizaciju određenih procesa, promjenu radnih navika, te izrazito utiče na kompetitivnost i produktivnost.

Kroz univerzitetsko obrazovanje, po uzoru na sve vodeće univerzitete u svijetu potrebno je formirati Master programe posebno fokusirane na AI. Kroz strateški pristup potrebno je definisati realističan cilj novih doktorskih i magistarskih pozicija iz ovih oblasti, da bi se proaktivno zadovoljile buduće potrebe tržišta rada. Pored specijaliziranih smjerova iz AI, radi širokog uticaja, važno je razmotriti potrebu uspostavljanja interdisciplinarnih programa sa AI temama kao dopunskim u okviru drugih programa (eng. minor). Ovo je izrazito bitno za podršku i usvajanje upotrebe AI u drugim oblastima (npr. pravo, psihologija, medicina itd.).

Obrazovni sistem treba da redovno unaprijeđuje kurikulum prema trenutnim dostignućima. Da bi se održala usklađenost sa razvojem u drugim akademskim institucijama (npr. u inostranstvu) kao i sa industrijom, potrebno je stimulisati prakse i mobilnost studenata i osoblja univerziteta kroz međunarodnu saradnju. Zbog izražene kompetitivnosti u AI oblasti, posebno je bitno stimulisanje vrhunskih rezultata i razvijanje liderstva u naučno-tehnološkom razvoju.

Pored formalnog obrazovanja, potrebno je ponuditi adekvatno neformalno obrazovanje pristupačno zainteresovanima za cjeloživotno učenje. Zadnjih godina trend online kurseva je veoma prisutan i nudi mnoštvo opcija iz ove oblasti. Posebno zanimljiv je primjer njemačke platforme KI-campus namjenski fokusirane na AI.  Po uzoru na ove platforme potrebno je razmotriti potrebu razvoja sličnih sadržaja za naše govorno područje. Pored toga, veoma je važna podrška organizovanju i učešću u konferencijama, radionicama i lokalnim, regionalnim i internacionalnim skupovima.

Pored stučne javnosti, potrebno je posebnu pažnju posvetiti edukaciji šire javnosti koja je krajnji korisnik umjetne inteligencije. To može uključivati posebne aktivnosti za formiranje javnog mijenja, kao što su javni demo događaji (npr. demonstriranje autonomnih taksi-vozila na određenim dionicama), tematske emisije na TV itd. Potrebno je da se javnost obrazuje za odgovornu upotrebu i samozaštitu, upozna sa pravima, te prevlada eventualne barijere. Ovo je posebno važno imajući u vidu mogućnosti AI (npr. kreiranje lažnih vijesti, lažnih videa generativnim modelima, itd.). Također je potrebno edukovati javnost za efikasnu upotrebu AI radi automatizacije određenih procesa, što u konačnici povećava produktivnost i održava kompetitivnost.

\subsubsection{Infrastruktura za istraživanje, razvoj i komercijalizaciju AI}
Zbog interdisciplinarne prirode, potrebe za računarskim resursima, velikim podacima, itd.,  AI istraživanje, razvoj i komercijalizacija zahtjevaju značajnu infrastrukturu. To zahtijeva značajna finansijska sredstva (npr. super-računari i podatkovni centri). Iako je u dugoročnim planovima potrebno planiranje izgradnje takve infrastrukture, već sada postoje mogućnosti ostvarivanja strateških partnerstava sa institucijama koje posjeduju potrebnu infrastrukturu. Određeni infrastrukturni preduslovi se ipak mogu lakše ostvariti, uz dosta manji napor, ali zahtjevaju proaktivan pristup. Primjer toga su dostupni podaci. Da bi AI bila primjenjiva u BiH, konkretni podatkovni setovi (i stalni izvori podataka) iz bh prakse su potrebni za razvoj i testiranje AI modela. U istaživačke svrhe, bh infrastrukturne kompanije mogu ponuditi uređene setove podataka (npr. set podataka o potrošnji električne energije za razvoj modela predikcije). Svakako, potrebno je kreirati politike koje bi stimulisale kompanije da čine dostupnim takve podatke kao i da zaštite privatnost korisnika.

\subsubsection{Uslovi za privlačenje i zadržavanje kompetitivnog ljudskog kapitala}
Iako se kroz edukaciju ljudski kapital razvija, podjednako je važno uspostaviti uslove za zadržavanje i dodatno privlačenje ljudskog kapitala. Iako je već duži period izražen trend centralizacije i okupljanja talenta na lokacije razvijenih centara, postoje određeni faktori koji ovaj trend usporavaju pa čak i okreću.

Trend omogućavanja rada na daljinu, freelance rada i digitalnih nomada predstavlja izvrsnu priliku za privlačenje regionalnog i internacionalnog ljudskog kapitala. Bosna i Hercegovina se nudi  kao primamljiva lokacija zbog relativno niskih troškova života, uređenosti, ugodne klime i raznovrsne prirode, relativno dobre povezanosti avionskim saobraćajem, vremenske zone usklađene sa mnogim razvijenim zemljama itd. Iako su ovakvi uslovi primamljivi za npr. bh dijasporu, postoje velike prepreke za strance koji ne poznaju lokalni jezik. Uslovi bi se vjerovatno mogli unaprijediti radom na poboljšanju avionske, saobraćajne i internet povezanosti, jasnije pravne regulative rada na daljinu (npr. porezni sistem i zdravstvo) i samog brendiranja i promocije BiH kao destinacije za rad na daljinu. Pored toga, važno je uspostaviti dinamično okruženje, sa sadržajem za razvoj, prilikama za kretanje (mobilnost), umrežavanje, posjetama i gostovanjima stručnjaka i konferencijama itd.

Neke zemlje imaju proaktivne pristupe za privlačenje uspostavljanja istraživačkih grupa velikih multi-nacionalnih korporacija. Poznato je da velike kompanije (npr. Google, Amazon, itd.) imaju svoje istraživačke centre širom svijeta. Zemlje pružaju razne olakšice ali kompanije prvenstveno razvijaju centre na lokacijama gdje je izražena izvrsnost iz određenih oblasti (npr. Amazon i centar mašinske vizije u Grazu u Austriji (Amazon.com, Inc., 2016)). Na taj način vlasti pomažu zadržavanju lokalnog talenta i privlačenju stranog. Ovo je odlična prilika koju i BiH može iskoristiti. Pored privatnih kompanija, dosta nacionalnih programa drugih zemalja podrazumjeva internacijalnu saradnju kojoj treba da se teži.

\subsubsection{Uslovi i prilike za primjene AI}
Kao krajnji rezultatat razvoja AI tehnologija, sama primjena kroz komercijalizaciju zahtjeva određene uslove. Pored prethodno navedenih infrastrukturnih uslova, prvenstveno su tu finansijski uslovi kroz potrebne investicije, pravni uslovi kroz regulative te uslovi tržišta kroz prihvatanje krajnjih korisnika i globalnu konkurentnost.

Za razliku od nekih drugih biznis modela, AI-bazirani biznis modeli zahtjevaju viši nivo inovacije i ulaganja u istraživanje. Zbog toga se u razvijenim zemljama često teži kolaboraciji industrije i akademije, te kompanije nude industrijske doktorske pozicije. Doktorske pozicije i u industriji i u akademiji predstavljaju rad na puno radno vrijeme. U mnogim aplikacijama intelektualna svojina treba biti adekvatno zaštićena patentima. Situacija u BiH vezano za ovo se može unaprijediti kroz finansijsko pomaganje istraživanja (npr. poreske olakšice za istraživanje, matching finansiranje ulaganja), poticanje spin-off kompanija na univerzitetima te strateško fokusiranje na dovođenje i ostvarivanje saradnje sa velikim kompanijama.

Zbog raznih barijera, mnogi problemi vezani za primjenu AI ne mogu biti riješeni od strane jedne kompanije (npr. nedostatak industrijskih standarda, prihvatljivost korisnika, itd.). S toga različiti programi saradnje i kooperacijski projekti finansirani od vladinih institucija pomažu zajedničkom otklanjanju tih barijera. 
Pravna regulativa, regulacija zakoni i industrijski standardi omogućavaju da se AI upotrebljava legalno i etično, te onemogući zloupotreba. Važno je da su ove norme u skladu sa tehničkim mogućnostima AI tehnologije te da se pri izdradi uključe mišljenja struke. EU ima već uspostavljene norme kojima se treba težiti. 

Kao što je predstavljeno AI predstavlja dinamičnu oblast sa stalnom potrebom za učenjem i istraživanjem da bi se održala kompetitivnost. Mnoge zemlje ulažu značajne napore da bi promovisale i podržale svoje kompanije da zadrže kompetitivnost na globalnom tržištu. Bitno je kreirati strateške oblasti gdje se može ostvariti značajna kompetitivna prednost i motivisati bh kompanije da djeluju u toj oblasti. Također je potrebno obezbijediti adekvatnu zastupljenost na relevantnim promotivnim događajima kao što su sajmovi i konferencije.

Pored toga, kao što je pomenuto i u oblasti edukacije, za dinamičan razvoj AI, veoma je važno imati i korisnike koji prate tok, aktivno koriste i vraćaju relevantne povratne infomacije o upotrebljivosti tehnologija. Pored prethodno pomenute edukacije, korisnike je potrebno animirati kao aktivne učesnike u AI revoluciji.

\section{Zaključak}
Kroz ovaj rad pokušali smo kompleksnu tematiku umjetne inteligencije (eng. Artificial Intelligence, AI) učiniti pristupačnijom široj publici i relevantnim učesnicima u odlučivanju. Nakon kratkog uvoda u samu oblast AI, primjena i utjecaja na društvo i ekonomiju, predstavili smo odabrane primjere aktualnih istraživanja i razvoja AI bh istraživača i stručnjaka. Odabrani primjeri uključuju različite aspekte AI, od razvoja metoda do primjena u različitim oblastima. Nakon toga smo predstavili i kratki osvrt na svjetske trendove. Kao poseban trend, izvojen je primjer izrade nacionalnih strategija te su posebno tretirane nacionalne strategije nekih zemalja Evrope, Amerike, Australije i regiona.

Kao krunski doprinos, na osnovu prezentiranih analiza i formiranog stručnog mišljenja, definisali smo viziju Bosne i Hercegovine u doba AI. Vizija je ambiciozna ali i realistična, prikladna bosansko-hercegovačkoj realnosti ali i nužnosti za djelovanjem u ovom krucijalnom momentu u ljudskom tehnološkom razvoju. Uz viziju, predstavljene su prioritetne oblasti za djelovanje i neki primjeri mogućih akcija. Istinski se nadamo da će formiranje ove vizije doprinijeti definisanju strateškog pristupa tretiranja tematike AI u BiH i daljim koracima ka pozicioniranju BiH na svjetskoj AI sceni.

\section*{Zahvala}
Autori su članovi Asocijacije za napredak nauke i tehnologije (ANNT). ANNT je nevladina organizacija koju čine mladi naučnici u oblastima prirodno-matematičkih i tehničkih nauka. Naši članovi žive i djeluju u Bosni i Hercegovini, dok dio njih radi na prestižnim svjetskim institucijama, ali su svi vezani za našu domovinu i dijele zajedničku viziju o unapređenju stanja nauke u njoj. Autori se također zahvaljuju dr. med. Merimi Pindžo za kritički pregled i stručne sugestije prilikom pisanja dijelova ovog rada. Istraživanje Zlatana Ajanovića je djelomično finansirano od strane European Research Council Starting Grant - TERI “Teaching Robots Interactively”, referenca projekta 804907.

\section*{Biografije}

Dr. \textbf{Zlatan Ajanović} je postdoktorski istraživač na Tehničkom Univerzitetu Delft, te eksterni predavač na Univerzitetu Primjenjenih Nauka Technikum-Wien u Beču. Bavi se istraživanjem metoda interaktivnog učenja i AI planiranja u robotici. Karijeru je počeo u Prevent Group u Bosni i Hercegovini gdje je radio na više projekata,  kao što je pokretanje novog proizvodnog procesa te korištenje kompjuterske vizije za osiguravanje kvalitete proizvoda. Dobitnik je najprestižnije evropske stipendije za PhD studij (Marie Skłodowska-Curie Fellowship), u sklopu ITEAM projekta. Kroz ITEAM projekat je bio zaposlen kao senior istraživač u istraživačkom centru Virtual Vehicle u Gracu, te kao gostujući istraživač na Tehničkom Univerzitetu Delft, Univerzitetu u Sarajevu, AVL List i Volvo Cars. Pored toga, učestvovao je u više internacionalnih projekata sa preko 50 partnera širom svijeta, te je aktivno učestvovao i vodio pripremanje uspješnih prijedloga projekata ukupne vrijednosti od preko 50 miliona eura. Bachelor i Master studije je završio na Elektrotehničkom fakultetu Univerziteta u Sarajevu u oblasti automatskog upravljanja. Titulu doktora tehničkih nauka stekao je na Tehničkom Univerzitetu u Grazu. Redovno objavljuje radove i prisustvuje na najprestižnijim događajima iz Robotike, Umjetne inteligencije i Automatskog upravljanja. Član je tehničkog komiteta za Inteligetna Autonomna Vozila i tehničkog komiteta za Inteligentno upravljanje pri IFAC te recenzent više časopisa i konferencija iz oblasti istraživanja. Za svoj rad, nagrađen je sa IFAC nagradom za najboljeg mladog autora te stipendijom Hans List Fond za svoju doktorsku disetraciju.

Dr. \textbf{Emina Aličković} je zaposlena kao vodeći naučnik u Eriksholm istraživačkom centru, koji je dio vodećeg svjetskog proizvođača slušnih aparata Oticon A/S, Danska. Također ima poziciju i vanrednog profesora na Odsjeku za elektrotehniku na Linkoping Univerzitetu u Švedskoj. Bachelor studije je završila na Internacionalnom Univerzitetu u Sarajevu, Odsjek za elektrotehniku i diplomirala je kao najbolji student generacije 2010 na Univerzitetu. Magistarske i doktorske studije je završila na Internacionalnom Burch Univerzitetu u Sarajevu, Odsjek za elektrotehniku i 2015. godine je stekla titulu doktora elektrotehničkih nauka. Autor je više desetina radova objavljenih u prestižnim naučnim časopisima i međunarodnim konferencijama i izumitelj je nekoliko međunarodnih patenata. Trenutno je mentor više doktoranata i postdoktoranata. Njeni istraživački interesi su statistička i adaptivna obrada signala i mašinsko učenje, sa primjenom u neuroznanosti i neurotehnologiji.

Dr. \textbf{Aida Branković} je završila Bachelor i Master studije na Elektrotehničkom fakultetu Univerziteta u Sarajevu (UNSA) 2009. i 2011. 2018. je dobila titulu doktora nauka informacionih tehnologija na Politekniku di Milano. Po završetku doktorada nastavlja kao istraživač nakon čega se pridružuje EMAGIN timu Kvinslend univerziteta i radi kao predavač ITEE odsjeku. Autor je 2 patenta, 1 simulatora, Svite algoritama “REPORT: Machine Learning driven real time clinical decision support tools”, te 16 naučnih radova objavljanjenih i prezentovanim u prestižnim časposima i konferencijama. Dobitnik je prestižnog granta valde “Advance Queensland Research Fellowship (AQRF)” 2020. Djeluje u polju nelinerane identifikacije, teoretskog mašinskog učenja i primjeni istih u bio-medicinskim domenu. Trenutno radi ako istraživač u Health Intelligence timu pri Naučnoj i industrijskoj istraživačkoj organizaciji Komonvelta (The Commonwealth Scientific and Industrial Research Organisation (CSIRO)) i spoljnji saradnik na Kvinslend univerzitetu, Australija.

Dr. \textbf{Sead Delalić} završio je prvi i drugi ciklus studija na Odsjeku za matematiku Prirodno-matematičkog fakulteta Univerziteta u Sarajevu (UNSA) sa najvećim prosjekom ocjena 10.0. Osvajač je dvije Zlatne značke Univerziteta kao najbolji student UNSA za 2015. i 2017. godinu.  Završio je treći ciklus studija “Matematičke nauke u jugoistočnoj Evropi” na UNSA, te je stekao zvanje doktor kompjuterskih nauka. Trenutno je zaposlen kao viši asistent na Prirodno-matematičkom fakultetu UNSA. Više od pet godina intenzivno sarađuje na industrijskim projektima u AI/ML odjelu kompanije Info Studio d.o.o., te od 2022. godine kao AI inžinjer u kompaniji Infobip. Fokus rada i istraživanja je primjena AI tehnika za poboljšanje postojećih platformi i web sistema, sa posebnim fokusom na oblasti optimizacija, metaheuristika, te mašinskog učenja i neuralnih mreža. Autor je više od 25 naučnih radova i jednog univerzitetskog udžbenika. Jedan je od osnivača Asocijacije za napredak nauke i tehnologije ANNT, te redovno učestvuje u pripremama najboljih BH mladih matematičara za međunarodna takmičenja.

\textbf{Eldar Kurtić} je Bachelor i Master studij završio na Elektrotehničkom fakultetu Univerziteta u Sarajevu (UNSA) sa odlikovanjem Zlatna značka Univerziteta u Sarajevu na I i II ciklusu studija. Po završetku studija radio je u istraživačkom centru Robert Bosch GmbH Zentrum für Forschung und Vorausentwicklung (Renningen, Germany) na razvoju robotskih manipulatora i mašinske vizije za automatsku detekciju kvarova na industrijskim proizvodima. Nakon toga, radio je u istraživačkom centru Virtual Vehicle (Graz, Austria) na razvoju algoritama planiranja kretanja i računarske vizije za autonomna vozila. Trenutno radi kao istraživač i softver inženjer u oblasti mašinskog učenja i umjetne inteligencije na Institute of Science and Technology Austria (Vienna, Austria), sa fokusom na kompresiji dubokih neuronskih mreža.

Dr. \textbf{Salem Malikić} je trenutno zaposlen kao istraživač na Nacionalnom institutu za karcinom u Bethesdi, savezna država Maryland, Sjedinjene Američke Države. Fakultetsko obrazovanje započeo je na Univerzitetu u Sarajevu gdje je završio Bachelor studij Teorijske kompjuterske nauke na Odsjeku za matematiku Prirodno-matematičkog fakulteta kao jedan od najboljih studenata univerziteta i dobitnik Zlatne značke. Nakon toga, školovanje nastavlja u Kanadi na Simon Fraser univerzitetu na kom je završio magistarski, a nakon toga i doktorski studij iz oblasti Bioinformatike. Za vrijeme doktorskih studija također je radio i kao gost istraživač na Federalnom institutu za tehnologiju u Cirihu (Švicarska), Indiana univerzitetu (Sjedinjene Američke Države) i Vankuverskom centru za prostatu (Kanada). Dobitnik je najprestižnije kanadske stipendije za doktorski studij (Vanier Canada Graduate Scholarship). Njegova istraživanja su fokusirana na razvoj algoritama za analizu nastanka i evolucije tumora i autor je 15 naučnih radova objavljenih u prestižnim žurnalima (uključujući Nature i Cell). Metodi koje je dizajnirao zajedno sa koautorima su korišteni za analizu podataka u nekim od najpoznatijih studija evolucije karcinoma. Pored primarnih istraživanja, dr. Malikić aktivno prati i matematička takmičenja u Bosni i Hercegovini i svijetu. Kao učenik osnovne i srednje škole predstavljao je Bosnu i Hercegovinu na pet međunarodnih olimpijada sa kojih se vratio sa četiri osvojene medalje. Autor je više članaka i problema iz domena olimpijske matematike objavljenih u nekoliko žurnala u Kanadi i Sjedinjenim Američkim Državama.

Dr. \textbf{Adnan Mehonić} je profesor na University College-u u Londonu (UCL), na Fakultetu za elektroniku i elektroinženjering, te naučni član Kraljevske akademije za tehničke nauke Velike Britanije, direktor magistarskog programa za nanotehonolgiju, suosnivač i tehnološki direktor “Intrinsic Semiconductor Technologies”, kompanije za razvoj čipova i sistema umjetne inteligencije. Autor je više od 100 radova objavljenih u prestižnim naučnim časopisima (uključujući časopis Nature) i međunarodnim konferencijama, inovator je 12 međunarodnih patenata. MIT Technology Review ga je uvrstio na listu od 35 svjetskih inovatora mlađih of 35 godina.

M.Sc. \textbf{Hamza Merzić} je senior istraživač u oblasti umjetne inteligencije u Google DeepMind-u u Londonu i jedan od osnivača Asocijacije za napredak nauke i tehnologije. Završio je Bachelor na Odsjeku za automatiku i elektroniku pri Elektrotehničkom fakultetu Univerziteta u Sarajevu (UNSA) 2015. godine kao najbolji student i dobitnik Zlatne značke Univerziteta. Tokom Bachelor studija obnašao je ulogu podpredsjednika IEEE Studentskog ogranka u Sarajevu u sklopu čega je radio na projektima poput razvoja autonomnog fotonaponskog sistema (solarno drvo) te organizovao radionice robotike za srednjoškolce i studente. Magistarski studij u oblasti robotike i umjetne inteligencije završio je 2018. godine na ETH Cirih u Švicarskoj u saradnji sa Max-Planck Institutom za inteligentne sisteme u Tubingenu, Njemačka. Tokom Master studija stekao je i iskustva u akademiji, radeći kao istraživač u ADRL grupi pri ETH Cirih gdje je radio na projektu automne izgradnje (kolaboracija ADRL i fakulteta za arhitekturu), te u start-up sferi radeći za Rapyuta Robotics, švicarsko-japansku firmu za autonomnu robotiku. Redovno objavljuje naučne radove iz oblasti robotike i umjetne inteligencije.

Dr. \textbf{Kenan Šehić} je uspješno odbranio doktorsku disertaciju iz oblasti računarske i primijenjene matematike na Tehničkom Univerzitetu u Danskoj 2020. godine. Trenutno je na postdoktorskom istraživanju na Lund Univerzitetu pri Odsjeku za računarsku nauku. Kao dobitnik MIT gostujuće stipendije, dio istraživačke karijere provodi na Massachusetts Institutu za Tehnologiju (MIT) pri Odsjeku za aeronautiku i astronautiku u grupi za kvantificiranje nesigurnosti. Alumni je Univerziteta u Sarajevu gdje završava prvi i drugi ciklus na Mašinskom fakultetu. Autor je 3 naučna rada iz oblasti kvantificiranje nesigurnosti i mašinskog učenje. Njegovi interesi su kvantificiranje nesigurnosti, mašinsko učenje i optimizacija hiperparametara. Član je Asocijacije za Napredak Nauke i Tehnologije.

\textbf{Bahrudin Trbalić} je doktorski kandidat Fizike na Stanford Univerzitetu. Proučava tamnu materiju koristeći direktne metode detekcije. Završio je studij fizike i elektrotehnike na MIT-u. Tokom svoje karijere je učestvovao u organizaciji mnogobrojnih kampova, predavanja i takmičenja iz oblasti fizike. Predstavnik je BiH na svjetskim olimpijadama iz fizike na kojima je osvojio 4 srebrene medalje.

\section*{References}

\medskip

About the JAIC - JAIC [WWW Document], 2019. URL https://www.ai.mil/about.html (accessed 5.1.22).

AiNed Foundation, 2021. Major interests and opportunities for the Netherlands with AI [WWW Document]. AiNed. URL https://ained.nl/en/ (accessed 5.1.22).

Ajanovic, Z., 2019. Towards SuperHuman autonomous vehicles. Graz University of Technology.

Ajanovic, Z., Lacevic, B., Shyrokau, B., Stolz, M., Horn, M., 2018. Search-Based Optimal Motion Planning for Automated Driving, in: 2018 IEEE/RSJ International Conference on Intelligent Robots and Systems (IROS). Presented at the 2018 IEEE/RSJ International Conference on Intelligent Robots and Systems (IROS), pp. 4523–4530. https://doi.org/10.1109/IROS.2018.8593813

Ajanovic, Z., Regolin, E., Horn, M., Ferrara, A., 2022. Search-Based Task and Motion Planning for Hybrid Systems: Aggressive Autonomous Vehicles. Engineering Applications of Artificial Intelligence (under revision).

Ajanovic, Z., Regolin, E., Stettinger, G., Horn, M., Ferrara, A., 2020. Search-based motion planning for performance autonomous driving, in: Klomp, M., Bruzelius, F., Nielsen, J., Hillemyr, A. (Eds.), Advances in Dynamics of Vehicles on Roads and Tracks. Springer International Publishing, Cham, pp. 1144–1154.

Alickovic, E., Kevric, J., Subasi, A., 2018. Performance evaluation of empirical mode decomposition, discrete wavelet transform, and wavelet packed decomposition for automated epileptic seizure detection and prediction. Biomedical signal processing and control 39, 94–102.

Alickovic, E., Lunner, T., Gustafsson, F., Ljung, L., 2019. A tutorial on auditory attention identification methods. Frontiers in neuroscience 153.

Alickovic, E., Lunner, T., Wendt, D., Fiedler, L., Hietkamp, R., Ng, E.H.N., Graversen, C., 2020. Neural representation enhanced for speech and reduced for background noise with a hearing aid noise reduction scheme during a selective attention task. Frontiers in neuroscience 846.

Alickovic, E., Ng, E.H.N., Fiedler, L., Santurette, S., Innes-Brown, H., Graversen, C., 2021. Effects of hearing aid noise reduction on early and late cortical representations of competing talkers in noise. Frontiers in Neuroscience 15.

Alickovic, E., Subasi, A., 2018. Ensemble SVM method for automatic sleep stage classification. IEEE Transactions on Instrumentation and Measurement 67, 1258–1265.
Amazon.com, Inc., 2016. New Austrian Development Center to support Prime Air [WWW Document]. EU About Amazon. URL https://www.aboutamazon.eu/news/press-lounge/new-austrian-development-center-to-support-prime-air (accessed 6.5.22).

Andersen, A.H., Santurette, S., Pedersen, M.S., Alickovic, E., Fiedler, L., Jensen, J., Behrens, T., 2021. Creating clarity in noisy environments by using deep learning in hearing aids, in: Seminars in Hearing. pp. 260–281.

Artificial Intelligence for the American People [WWW Document], 2021. URL https://trumpwhitehouse.archives.gov/ai/ai-american-innovation/ (accessed 5.1.22).

Austrian Council on Robotics and Artificial Intelligence, 2018. Shaping the Future of Austria with Robotics and Artificial Intelligence [WWW Document]. URL https://www.acrai.at/wp-content/uploads/2019/08/ACRAI\_White\_Paper\_EN.pdf (accessed 5.1.22).

Basu, K., Sinha, R., Ong, A., Basu, T., 2020. Artificial Intelligence: How is It Changing Medical Sciences and Its Future? Indian J Dermatol 65, 365–370. https://doi.org/10.4103/ijd.IJD\_421\_20

Bilert, s, Hjortkjær, J., Fuglsang, S., Alickovic, E., Shiell, M., Rotger-Griful, S., Zaar, J., 2020. Neural audio-visual speech processing during selective listening in a monologue vs. dialogue paradigm. Presented at the Advances and Perspectives in Auditory Neuroscience 2020.

Brown, T., Mann, B., Ryder, N., Subbiah, M., Kaplan, J.D., Dhariwal, P., Neelakantan, A., Shyam, P., Sastry, G., Askell, A., Agarwal, S., Herbert-Voss, A., Krueger, G., Henighan, T., Child, R., Ramesh, A., Ziegler, D., Wu, J., Winter, C., Hesse, C., Chen, M., Sigler, E., Litwin, M., Gray, S., Chess, B., Clark, J., Berner, C., McCandlish, S., Radford, A., Sutskever, I., Amodei, D., 2020. Language models are few-shot learners, in: Larochelle, H., Ranzato, M., Hadsell, R., Balcan, M.F., Lin, H. (Eds.), Advances in Neural Information Processing Systems. Curran Associates, Inc., pp. 1877–1901.
Campbell, M., Hoane, A.J., Hsu, F., 2002. Deep Blue. Artificial Intelligence 134, 57–83. https://doi.org/10.1016/S0004-3702(01)00129-1

Chowdhery, A., Narang, S., Devlin, J., Bosma, M., Mishra, G., Roberts, A., Barham, P., Chung, H.W., Sutton, C., Gehrmann, S., Schuh, P., Shi, K., Tsvyashchenko, S., Maynez, J., Rao, A., Barnes, P., Tay, Y., Shazeer, N., Prabhakaran, V., Reif, E., Du, N., Hutchinson, B., Pope, R., Bradbury, J., Austin, J., Isard, M., Gur-Ari, G., Yin, P., Duke, T., Levskaya, A., Ghemawat, S., Dev, S., Michalewski, H., Garcia, X., Misra, V., Robinson, K., Fedus, L., Zhou, D., Ippolito, D., Luan, D., Lim, H., Zoph, B., Spiridonov, A., Sepassi, R., Dohan, D., Agrawal, S., Omernick, M., Dai, A.M., Pillai, T.S., Pellat, M., Lewkowycz, A., Moreira, E., Child, R., Polozov, O., Lee, K., Zhou, Z., Wang, X., Saeta, B., Diaz, M., Firat, O., Catasta, M., Wei, J., Meier-Hellstern, K., Eck, D., Dean, J., Petrov, S., Fiedel, N., 2022. PaLM: Scaling Language Modeling with Pathways (No. arXiv:2204.02311). arXiv. https://doi.org/10.48550/arXiv.2204.02311

Delalić, S., Alihodžić, A., Tuba, M., Selmanović, E., Hasić, D., 2020. Discrete bat algorithm for event planning optimization, in: 2020 43rd International Convention on Information, Communication and Electronic Technology (MIPRO). pp. 1085–1090.

Deng, J., Dong, W., Socher, R., Li, L.-J., Li, K., Fei-Fei, L., 2009. Imagenet: A large-scale hierarchical image database, in: 2009 IEEE Conference on Computer Vision and Pattern Recognition. pp. 248–255.

Department of Industry, S., 2021. Australia’s Artificial Intelligence Action Plan [WWW Document]. Department of Industry, Science, Energy and Resources. URL https://www.industry.gov.au/data-and-publications/australias-artificial-intelligence-action-plan (accessed 5.1.22).

Does AI have a hardware problem?, 2018. . Nat Electron 1, 205–205. https://doi.org/10.1038/s41928-018-0068-2
EC/OECD, 2021. AI Strategies and Policies in Netherlands [WWW Document]. URL https://oecd.ai/en/dashboards/countries/Netherlands (accessed 5.1.22).

Federal Ministry Republic of Austria for Digital and Economic Affairs (BMDW), 2021. Strategie der Bundesregierung für Künstliche Intelligenz - Artificial Intelligence Mission Austria 2030 (AIM AT2030) [WWW Document]. URL https://www.bmdw.gv.at/en/Topics/Digitalisation/Strategy/Artificial-Intelligence.html (accessed 5.1.22).

Garrett, C.R., Chitnis, R., Holladay, R., Kim, B., Silver, T., Kaelbling, L.P., Lozano-Pérez, T., 2021. Integrated task and motion planning. Annual review of control, robotics, and autonomous systems 4, 265–293.

Geirnaert, S., Vandecappelle, S., Alickovic, E., de Cheveigne, A., Lalor, E., Meyer, B.T., Miran, S., Francart, T., Bertrand, A., 2021. Electroencephalography-based auditory attention decoding: Toward neurosteered hearing devices. IEEE Signal Processing Magazine 38, 89–102.

Government Offices of Sweden, 2018. National approach to artificial intelligence [WWW Document]. URL https://www.government.se/491fa7/contentassets/fe2ba005fb49433587574c513a837fac/national-approach-toartificial-intelligence.pdf (accessed 5.1.22).

Hajkowicz, S., Karimi, S., Wark, T., Chen, C., Evans, M., Rens, N., Dawson, D., Charlton, A., Brennan, T., Moffatt, C., others, 2019. Artificial intelligence: Solving problems, growing the economy and improving our quality of life.

Hall, W., Pesenti, J., 2017. Growing the artificial intelligence industry in the UK. Department for Digital, Culture, Media \& Sport and Department for Business, Energy \& Industrial Strategy. Part of the Industrial Strategy UK and the Commonwealth.

Hrvatski sabor, 2021. Nacionalna razvojna strategija Republike Hrvatske do 2030. godine [WWW Document]. URL https://narodne-novine.nn.hr/clanci/sluzbeni/2021\_02\_13\_230.html (accessed 6.3.22).

IFR, 2020. World Robotics Report 2020 [WWW Document]. IFR International Federation of Robotics. URL https://ifr.org/ifr-press-releases/news/record-2.7-million-robots-work-in-factories-around-the-globe (accessed 5.31.22).

Jacobs, C., van Ginneken, B., 2019. Google’s lung cancer AI: a promising tool that needs further validation. Nat Rev Clin Oncol 16, 532–533. https://doi.org/10.1038/s41571-019-0248-7

Karamehmedović, M., Šehić, K., Dammann, B., Suljagić, M., Karamehmedović, E., 2019. Autoencoder-aided measurement of concentration from a single line of speckle. Optics Express 27, 29098–29123.

Kober, J., Bagnell, J.A., Peters, J., 2013. Reinforcement learning in robotics: A survey. The International Journal of Robotics Research 32, 1238–1274.

Krizhevsky, A., Sutskever, I., Hinton, G.E., 2012. Imagenet classification with deep convolutional neural networks. Advances in neural information processing systems 25.

Kurtic, E., Campos, D., Nguyen, T., Frantar, E., Kurtz, M., Fineran, B., ... \& Alistarh, D. (2022). The Optimal BERT Surgeon: Scalable and Accurate Second-Order Pruning for Large Language Models. arXiv preprint arXiv:2203.07259.

Maintaining American Leadership in Artificial Intelligence [WWW Document], 2019. . Federal Register. URL https://www.federalregister.gov/documents/2019/02/14/2019-02544/maintaining-american-leadership-in-artificial-intelligence (accessed 5.1.22).

McCarthy, J., 2007. What is artificial intelligence?

McCarthy, J., 1958. Programs with common sense. RLE and MIT computation center Cambridge, MA, USA.

McCarthy, J., Minsky, M.L., Rochester, N., Shannon, C.E., 2006. A Proposal for the Dartmouth Summer Research Project on Artificial Intelligence, August 31, 1955. AI Magazine 27, 12–12. https://doi.org/10.1609/aimag.v27i4.1904

Mehonic, A., Kenyon, A.J., 2022. Brain-inspired computing needs a master plan. Nature 604, 255–260.

Ministry of Finance and Ministry of Industry, Business and, Financial Affairs of Denmark, 2019. National Strategy for Artificial Intelligence.

National Artificial Intelligence R\&D Strategic Plan -- OSTP invites input for update (by 3/4) - EconSpark [WWW Document], 2022. URL https://www-aeaweb-org.tudelft.idm.oclc.org/forum/2341/national-artificial-intelligence-strategic-invites-update (accessed 5.1.22).

Newell, A., Simon, H., 1956. The logic theory machine–A complex information processing system. IRE Transactions on Information Theory 2, 61–79. https://doi.org/10.1109/TIT.1956.1056797

Ng, A., 2018. AI is the new electricity. O’Reilly Media.

NL AI Coalition, 2022. Algorithms that work for everyone | NL AIC [WWW Document]. Nederlandse AI Coalitie. URL https://nlaic.com/en/ (accessed 5.1.22).

Osaba, E., Villar-Rodriguez, E., Del Ser, J., Nebro, A.J., Molina, D., LaTorre, A., Suganthan, P.N., Coello, C.A.C., Herrera, F., 2021. A Tutorial On the design, experimentation and application of metaheuristic algorithms to real-World optimization problems. Swarm and Evolutionary Computation 64, 100888.

Pentagon to spend \$874 million on artificial intelligence (AI) and machine learning technologies next year | Military Aerospace [WWW Document], 2021. URL https://www.militaryaerospace.com/computers/article/14204595/artificial-intelligence-ai-dod-budget-machine-learning (accessed 5.1.22).

Perez-Dattari, R., Celemin, C., Franzese, G., Ruiz-del-Solar, J., Kober, J., 2020. Interactive Learning of Temporal Features for Control: Shaping Policies and State Representations From Human Feedback. IEEE Robotics \& Automation Magazine 27, 46–54. https://doi.org/10.1109/MRA.2020.2983649

PricewaterhouseCoopers, 2018. AI will create as many jobs as it displaces by boosting economic growth [WWW Document]. PwC. URL https://www.pwc.co.uk/press-room/press-releases/AI-will-create-as-many-jobs-as-it-displaces-by-boosting-economic-growth.html (accessed 6.6.22).

Raissi, M., Perdikaris, P., Karniadakis, G.E., 2017. Physics Informed Deep Learning (Part I): Data-driven Solutions of Nonlinear Partial Differential Equations (No. arXiv:1711.10561). arXiv. https://doi.org/10.48550/arXiv.1711.10561

Russell, S., Norvig, P., 2021. Artificial intelligence: A Modern Approach, 4th ed.

Sani, O.G., Yang, Y., Lee, M.B., Dawes, H.E., Chang, E.F., Shanechi, M.M., 2018. Mood variations decoded from multi-site intracranial human brain activity. Nature biotechnology 36, 954–961.

Science, N., Intelligence, T.C. (US). S.C. on A., 2019. The national artificial intelligence research and development strategic plan: 2019 update. National Science and Technology Council (US), Select Committee on Artificial ….

Šehić, K., Gramfort, A., Salmon, J., Nardi, L., 2021. LassoBench: A high-dimensional hyperparameter optimization benchmark suite for lasso. arXiv preprint arXiv:2111.02790.
Shahriari, B., Swersky, K., Wang, Z., Adams, R.P., De Freitas, N., 2015. Taking the human out of the loop: A review of Bayesian optimization. Proceedings of the IEEE 104, 148–175.

„Službene novine Federacije BiH“, broj 40/22, 2021. Strategija razvoja Federacije Bosne i Hercegovine [WWW Document]. (accessed 5.1.22).

srbija.gov.rs, 2020. Strategy for the Development of Artificial Intelligence in the Republic of Serbia for the period 2020-2025 [WWW Document]. URL https://www.srbija.gov.rs/tekst/en/149169/strategy-for-the-development-of-artificial-intelligence-in-the-republic-of-serbia-for-the-period-2020-2025.php (accessed 6.3.22).

Strubell, E., Ganesh, A., McCallum, A., 2019. Energy and policy considerations for deep learning in NLP, in: Proceedings of the 57th Annual Meeting of the Association for Computational Linguistics. pp. 3645–3650.

The Netherlands, Ministry of Economic Affairs and Climate Change, 2019. Strategic Action Plan for Artificial Intelligence.

The Royal Society, 2019. The AI revolution in science: applications and new research directions | Royal Society [WWW Document]. URL http://royalsociety.org/blog/2019/08/the-ai-revolution-in-science/ (accessed 5.1.22).

Turing, A., 1950. Computing machinery and intelligence.

Vinnova, R., 2018. Artificial intelligence in Swedish business and society: analysis of development and potential.

Žunić, E., Delalić, S., Đonko, D., 2022. Adaptive multi-phase approach for solving the realistic vehicle routing problems in logistics with innovative comparison method for evaluation based on real GPS data. Transportation Letters 14, 143–156.

Žunić, E., Hodžić, K., Delalić, S., Hasić, H., Handfield, R.B., 2021. Application of data science in supply chain management: Real-world case study in logistics, in: Data Science and Its Applications. Chapman and Hall/CRC, pp. 205–237.

\end{document}